\documentclass{bmvc2k}
\usepackage{hyperref}
\usepackage{amsfonts}

\usepackage{orcidlink}
\usepackage{colortbl}  % Allows coloring of table cells
\definecolor{lightgreen}{rgb}{0.8,1,0.8}
\usepackage{booktabs}
\usepackage{subfigure}
\usepackage{subcaption}
\title{Introducing \texttt{SDICE}: An Index for Assessing Diversity of Synthetic Medical Datasets}

\addauthor{Mohammed Talha Alam}{mohammed.alam@mbzuai.ac.ae}{1}
\addauthor{Raza Imam}{raza.imam@mbzuai.ac.ae}{1}
\addauthor{Mohammad Areeb Qazi}{mohammad.qazi@mbzuai.ac.ae}{1}
\addauthor{Asim Ukaye}{asim.ukaye@mbzuai.ac.ae}{1}
\addauthor{Karthik Nandakumar}{karthik.nandakumar@mbzuai.ac.ae}{2}

\addinstitution{
  Mohamed bin Zayed University of
Artificial Intelligence,\\
 Abu Dhabi,\\
 United Arab Emirates
}
\runninghead{Alam et al}{SDICE Index}

\begin{document}

\maketitle

\begin{abstract}
  Advancements in generative modeling are pushing the state-of-the-art in synthetic medical image generation. These synthetic images can serve as an effective data augmentation method to aid the development of more accurate machine learning models for medical image analysis. While the fidelity of these synthetic images has progressively increased, the diversity of these images is an understudied phenomenon. In this work, we propose the \texttt{SDICE} index, which is based on the characterization of \underline{s}imilarity \underline{d}istributions \underline{i}nduced by a \underline{c}ontrastive \underline{e}ncoder. Given a synthetic dataset and a reference dataset of real images, the \texttt{SDICE} index measures the distance between the similarity score distributions of original and synthetic images, where the similarity scores are estimated using a pre-trained contrastive encoder. This distance is then normalized using an exponential function to provide a consistent metric that can be easily compared across domains. Experiments conducted on the MIMIC-chest X-ray and ImageNet datasets demonstrate the effectiveness of \texttt{SDICE} index in assessing synthetic medical dataset diversity.
  % \keywords{Diversity assessment \and Generative models \and Synthetic datasets \and Similarity distributions \and Contrastive encoder.}
\end{abstract}
\section{Introduction}
\label{sec:intro}

The limited size of medical imaging datasets is often a major roadblock in the development of accurate deep neural network (DNN) models for such domains. While datasets like ImageNet \cite{deng2009imagenet}, MS-COCO \cite{lin2014microsoft}, and LAION-400M \cite{schuhmann2021laion400m} have been instrumental in the advancement of DNN models, the high costs and expertise required for medical image collection and annotation inhibit the curation of such large-scale medical datasets. Patient privacy concerns and strict regulations such as GDPR \cite{gdpr} and HIPAA \cite{hipaa} further impede the sharing of routine medical datasets within the research community. Latent Diffusion Models (LDMs) such as Stable Diffusion \cite{rombach2022high} generate high-fidelity synthetic images conditioned on text prompts. Several implementations of Stable Diffusion have been proposed in the medical domain including RoentGen \cite{chambon2022roentgen} for Chest X-ray (CXR) generation, Medical Diffusion \cite{khader2022medical}, and Brain Imaging Generation \cite{pinaya2022brain} for MRI and CT image generation. While these works show that high-fidelity synthetic images can be generated, \textit{the ability of these image generation tools to produce synthetic datasets that encompass possible real-world variations is questionable}.
\begin{figure}[t]
    \centering
    \includegraphics[width=0.63\textwidth]{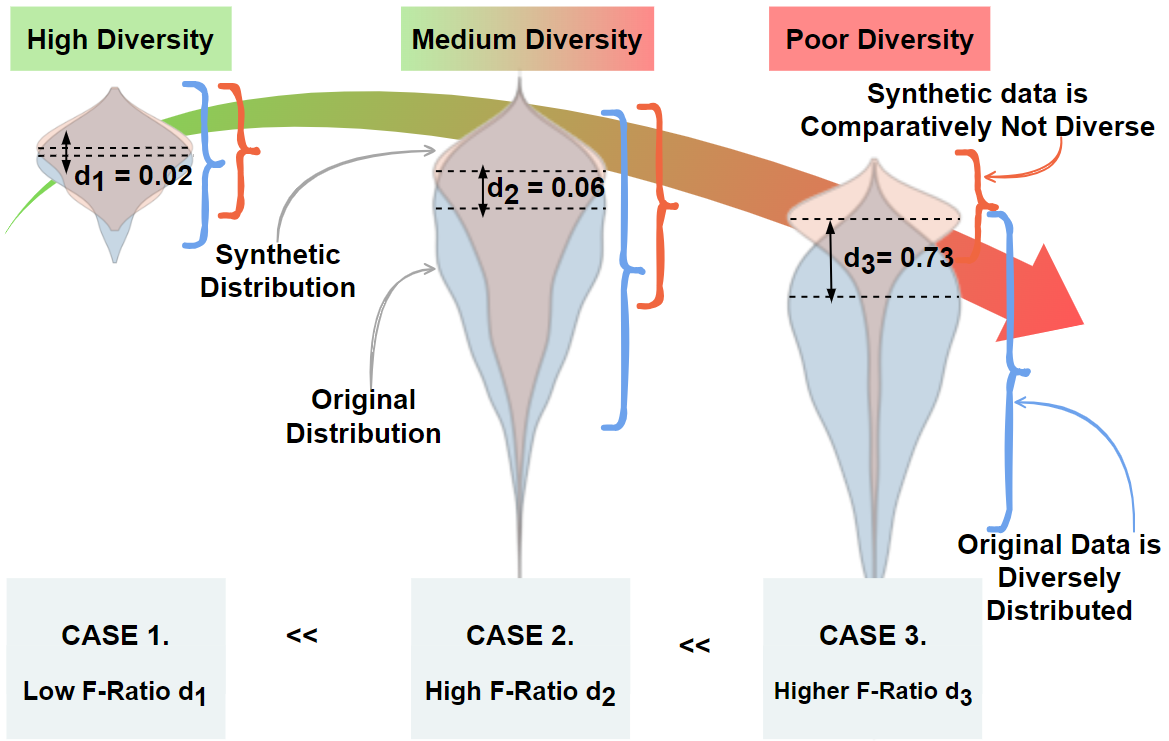}
    \vspace{10pt}
    \caption{F-ratio between the similarity score distribution of real and synthetic datasets serves as a good \textbf{indication} of the diversity within the synthetic dataset.}
    \label{fig:concept}
\end{figure}
Diversity of a synthetic dataset can be broadly defined as the spectrum of features, styles, and semantic variations contained in the generated images. Ensuring diversity in synthetic datasets is essential, as insufficient diversity can impair a model's generalization to real data \cite{jordon2022synthetic}. Diversity of synthetic datasets is typically evaluated using the Multi-Scale Structural Similarity Index (MS-SSIM) \cite{wang2003multiscale} score. A lower average MS-SSIM score is considered as a proxy for good dataset diversity. While the MS-SSIM score allows for an objective diversity assessment of synthetic datasets, it has inherent limitations. Firstly, it is computed at the image level and then extrapolated to the dataset by estimating first and/or second-order statistics. Secondly, it is typically not normalized, which implies that its absolute value is not very meaningful. However, it is still useful for relative comparisons between two competing datasets or methods. \cite{chambon2022roentgen} shows that the MS-SSIM is a poor indicator of diversity in CXR generation and yields inconclusive correlations. It must be emphasized that the MS-SSIM score was originally formulated as an analytical way to assess the quality of digital pictures by assessing structural similarity.\\
% Further, a good indicator of diversity is necessary to detect and avoid instances of training data memorization \cite{akbar2023beware}.
% \cite{naeem2020reliable} show that FID and precision/recall metrics are unreliable as they fail to detect the match between two identical distributions and are not robust against outliers. 
% Saad et al. \cite{saad2023assessing} show high variance in results when assessing the intra-class diversity of generated images in medical and non-biomedical using MS-SSIM and cosine distance. 
%Authors of RoentGen   
% Carlini et al. \cite{carlini2023extracting} show that Diffusion Models are susceptible to memorization and vulnerable to leaking training data. Akbar et al. \cite{akbar2023beware} study the instances of memorization in diffusion models applied to medical images.
% In this study, we address the need for an improved way to assess the diversity of a diffusion model. 

In this work, we propose a novel approach for diversity quantification of synthetic datasets.
% It must be emphasized that any meaningful quantification requires a baseline reference, and in the context of diversity assessment, the baseline is a dataset of real images. 
Given a sufficiently-diverse reference dataset of real images and a synthetic dataset, it is possible to analyze whether the variations in the synthetic dataset match or exceed those observed in the reference dataset, as shown in Figure \ref{fig:concept}. Specifically, we characterize the observed variations in a synthetic dataset by analyzing the similarity distributions between images of the same class (intra-class) and images from different classes (inter-class). We assume that the similarity scores are computed based on a contrastive encoder, which is pre-trained to be invariant under different affine/photometric transformations of the same image. We hypothesize that benchmarking of intra- and inter-class synthetic similarity distributions against their counterparts based on a reference dataset is a good proxy for diversity. Based on this hypothesis, we make the following contributions:
\begin{itemize}%[label=\textbf{--}]
    \item We propose a dataset-level diversity assessment index called \texttt{SDICE}, which characterizes \underline{S}imilarity \underline{Di}stributions \underline{I}nduced by a \underline{C}ontrastive \underline{E}ncoder. 
    \item While the concept of \texttt{SDICE} is generic, we also propose a specific instantiation of the \texttt{SDICE} index using F-ratio as the distance between two distributions and applying an exponential normalization to the resulting distance.
    \item We demonstrate the utility of the \texttt{SDICE} index by applying it to synthetic datasets generated from two models: (i) RoentGen trained on MIMIC-CXR (Chest X-ray) images and (ii) Stable Diffusion trained on natural images. Our analysis indicates that the generated synthetic CXR dataset has low diversity, especially failing to capture variations within the same class. 
\end{itemize}
\textbf{Related Work:} 
% \textbf{Latent Diffusion Models for Medical Image Generation}: Several works \cite{pinaya2022brain} use LDMs with and without conditions to generate synthetic images. Khader et al. \cite{khader2022medical} show that synthetic images improve performance in downstream tasks when there is limited data. Packhäuser et al. \cite{packhauser2023generation} have investigated the use of LDMs for generating CXR images, comparing them to GAN-based methods. Chambon et al. \cite{chambon2022adapting} show that fine-tuning Stable Diffusion \cite{rombach2022high} for generating high-fidelity CXR images for a single classification task. They expand on this work by proposing RoentGen \cite{chambon2022roentgen}, which improves the process and provides significant results on multiple classification tasks and report generation.\\
% \noindent \textbf{Evaluation of Generative Models}: 
% Image generation models are often evaluated based on the fidelity, diversity, and factual accuracy of their outputs. FID score is often used for evaluating the fidelity of LDMs \cite{rombach2022high}. In \cite{naeem2020reliable}, it was shown that FID and precision/recall metrics are unreliable; they claim these fail to detect the match between two identical distributions and are not robust against outliers. Hence, they propose density and coverage metrics to overcome these challenges. \cite{akbar2023beware} show that FID and Inception scores fail to capture instances of training data memorization in generative models. \\
\noindent Saad et al. \cite{saad2023assessing} show high variance in results when assessing the intra-class diversity of generated images in medical and non-biomedical domains using MS-SSIM and cosine distance. Friedman et. al. \cite{friedman2022vendi} argue that existing metrics for measuring diversity are often domain-specific and limited in flexibility and propose the Vendi score. They show that even models that capture all the modes of a labeled dataset can be less diverse than the real dataset. Alaa et al. \cite{alaa2022faithful} introduce a 3-d metric that characterizes the fidelity, diversity, and generalization performance of any generative model. They quantify diversity in the feature space, while our \texttt{SDICE} index operates in the similarity space. This distinction makes our method more robust, as it captures diversity across all clusters, unlike $\beta$-Recall, which is sensitive to the value of $\beta$ and may struggle with highly multimodal distributions.
% , where they judge the quality of individual samples generated by a (black-box) model after discarding low-quality samples.

\section{Proposed \texttt{SDICE} Index}
\label{sec:Methods}
\begin{figure*}[t]
    \centering
    \includegraphics[width=0.9\textwidth]{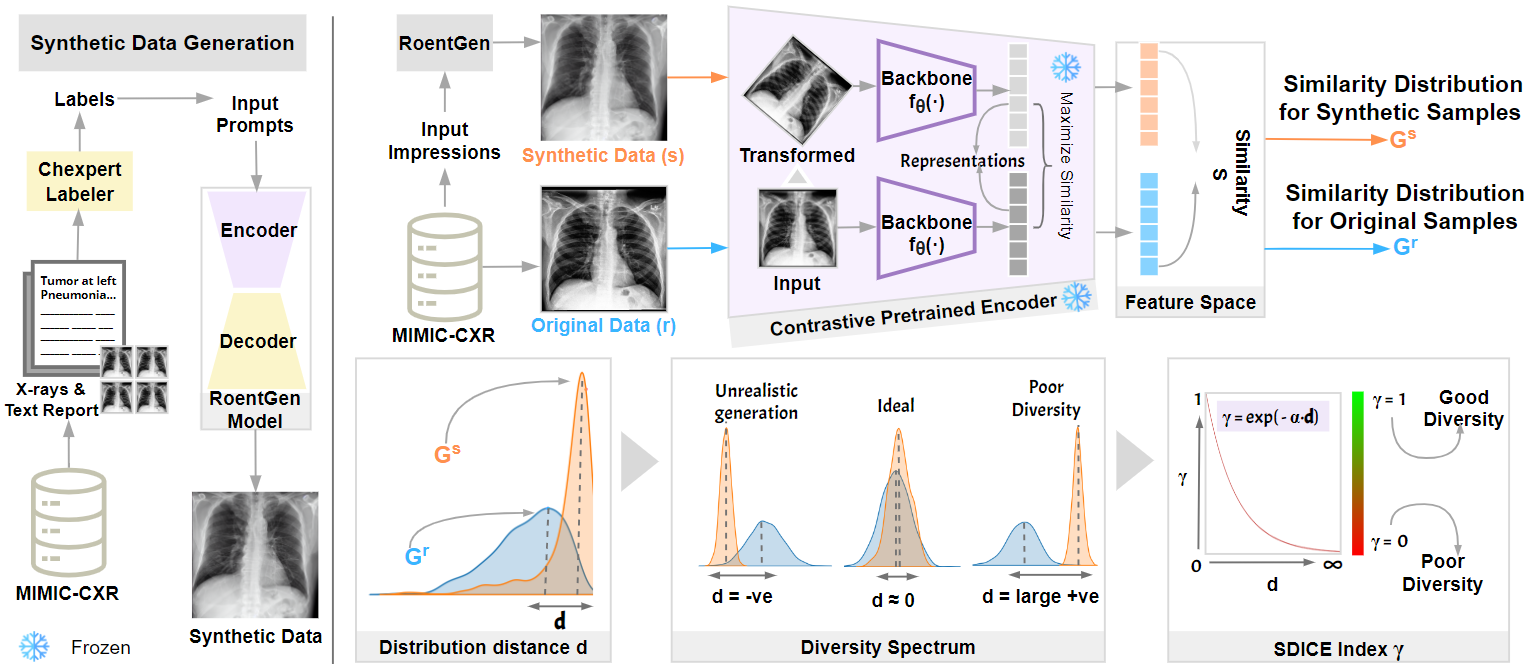}
    \vspace{8.9pt}
    \caption{
    \textbf{Overview of the proposed \texttt{SDICE} index.} We input the real and synthetic dataset to the contrastive pretrained encoder to obtain similarity score distributions. The F-ratio between the two distributions after exponential normalization can be used to assess the diversity of the synthetic dataset.}
    \label{fig:method}
\end{figure*}

The key intuition underlying the proposed \texttt{SDICE} index is that a synthetic dataset can be considered to have good diversity if the variations in this dataset closely follow or exceed the variations observed in a reference dataset containing sufficiently-diverse real images. Figure \ref{fig:method} provides an overview of the architecture of our proposed \texttt{SDICE} index. However, two main challenges need to be overcome to determine if two datasets (synthetic and real) have similar variations. 1) The variations in a dataset can be caused due to many reasons such as image noise and within and between class differences, and it is essential to capture these variations individually. 2) A good metric is required to capture pair-wise similarities between the images. \\
% Finally, we need a method to characterize the differences in the similarity score distributions. 

\noindent \textbf{Problem Statement}: Let, $\mathcal{D}^s = \{\mathbf{x}_i^s,y_i^s\}_{i=1}^{N^s}$ be a synthetic image dataset with $N^s$ samples, where 
$\mathbf{x} \in \mathcal{R}^{H \times W \times C}$
is an input image ($H$, $W$, and $C$ represent the height, width, and number of channels in the input image, respectively) and 
$y \in \{1,2,\cdots, M\}$ is the class label. Similarly, let $\mathcal{D}^r = \{\mathbf{x}_j^r,y_j^r\}_{j=1}^{N^r}$ be a real image dataset containing $N^r$ samples.
Let $\mathcal{F}: \mathcal{R}^{H \times W \times C} \rightarrow \mathcal{R}^{dim}$ be a pre-trained feature extractor that outputs a fixed-length embedding for a given image. Let $\mathcal{S}: \mathcal{R}^{dim} \times \mathcal{R}^{dim} \rightarrow \mathcal{R}$ be a similarity metric that outputs the similarity between two feature embeddings.
Given a synthetic dataset $\mathcal{D}^s$, a reference real dataset $\mathcal{D}^r$, a feature extractor $\mathcal{F}$, and a similarity metric $\mathcal{S}$, the goal is to compute a diversity index $\gamma \in [0,1]$ that indicates if the two datasets $\mathcal{D}^s$ and $\mathcal{D}^r$ have similar variations. A higher value of $\gamma$ indicates better diversity.

\subsection{Generic \texttt{SDICE} Index}
% The first step is to assess the variations in a given dataset, which could be categorized into three types:
The variations in a dataset can be broadly categorized into two types:

\begin{enumerate}
    \item \textbf{Intra-class Variations} ($intra$): These are variations between input images belonging to the same class, e.g., differences between CXR images of the same underlying condition. A well-trained feature extractor will produce embeddings that have high similarity for images of the same class.
    \item \textbf{Inter-class Variations} ($inter$): These are variations between input images belonging to different classes, e.g., differences between CXR images of patients having different diseases. A good feature extractor will learn to amplify these variations and produce embeddings that have lower similarity scores.
\end{enumerate}
To characterize the above types of variations, we employ the following approach. Let $\mathbf{x}_a^{*}$ and $\mathbf{x}_b^{*}$ be a pair of images randomly drawn from dataset $\mathcal{D}^{*}$, where $* \in \{s,r\}$. Let $S_{ab}^{*} = \mathcal{S}(\mathcal{F}(\mathbf{x}_a^{*}),\mathcal{F}(\mathbf{x}_b^{*}))$ be the similarity between two input images drawn from $\mathcal{D}^{*}$. Let $\mathcal{G}_{intra}^{*}$ be the probability distribution of $S_{ab}^{*}$ when $[y_a^* = y_b^*]$ (i.e., distribution of similarity scores between images of the same class). Similarly, let $\mathcal{G}_{inter}^{*}$ be the probability distribution of $S_{ab}^{*}$ when $[y_a^* \neq y_b^*]$ (i.e., distribution of similarity scores between images of different classes). Let $\mathbb{Q}:\mathcal{G} \times \mathcal{G} \rightarrow [0,\infty]$ be a distance measure between two probability distributions. Specifically, let $\mathbb{Q}(\mathcal{G}_0||\mathcal{G}_1)$ be the distance of a probability distribution $\mathcal{G}_0$ from another distribution $\mathcal{G}_1$. We can compute $d_{intra} = \mathbb{Q}(\mathcal{G}_{intra}^s||\mathcal{G}_{intra}^r)$ and $d_{inter} = \mathbb{Q}(\mathcal{G}_{inter}^s||\mathcal{G}_{inter}^r)$. Larger values of $d_{intra}$ ($d_{inter}$) indicate that the synthetic similarity distribution is highly dissimilar to the real similarity distribution, indicating low intra-class (inter-class) diversity. On the other hand, smaller values of $d_{intra}$  or $d_{inter}$ indicate good alignment between the score distributions, representing high diversity. Therefore, the diversity index $\gamma$ should be inversely proportional to the above distance values.
One limitation of the above distance values is their unbounded and unnormalized nature, which makes it difficult to interpret these values across domains. To address this issue, we introduce a normalization function $\mathcal{H}:[0,\infty] \rightarrow [0,1]$ with parameter $\alpha$ to obtain the \texttt{SDICE} index as:
\begin{equation}
\gamma_{intra} = \mathcal{H}_{\alpha}(d_{intra})
\end{equation}
\begin{equation}
\gamma_{inter} = \mathcal{H}_{\alpha}(d_{inter})
\end{equation}
The tuple $\texttt{SDICE}:=(\gamma_{intra},\gamma_{inter})$ can be used to assess the diversity of a synthetic dataset. If a single diversity index is desired, $\gamma$ can be defined as: 
\begin{equation}
\gamma = \sqrt{\gamma_{intra}^2+\gamma_{inter}^2}
\end{equation}
Note that higher values of $\gamma$ indicate better diversity.  

\subsection{Practical Implementation of \texttt{SDICE} Index}
% \begin{enumerate}
%     \item Feature Extractor $(\mathcal{F})$ 
%     \item Similarity Function $(\mathcal{S})$ 
%     \item Probability Distance Measure $(\mathbb{Q})$ and
%     \item Normalization Function $(\mathcal{H})$
% \end{enumerate} 
Four critical design choices must be made to practically implement the proposed \texttt{SDICE} index: (i) feature extractor $\mathcal{F}$, (ii) similarity function $\mathcal{S}$, (iii) probability distance measure $\mathbb{Q}$, and (iv) normalization function $\mathcal{H}$. Before making these design choices, one needs to understand the worst-case scenario for the diversity of a synthetic dataset. A synthetic dataset can be considered to have negligible diversity if all the generated images are either exact copies of each other or minor geometric and/or photometric variations of one another. In this worst-case scenario, the combination of feature extractor and similarity metric must result in a very high similarity value for any pair of images drawn from such a low diversity dataset. This can be achieved by training the feature extractor $\mathcal{F}$ in a self-supervised contrastive manner \cite{chen2020simple}, where different minor transformations (augmentations) of the same image are forced to produce identical feature vectors. This explains our choice of a pre-trained contrastive encoder for $\mathcal{F}$. Since cosine similarity is typically employed in such contrastive encoders, we also choose cosine similarity as the default similarity function.

One common approach to estimate the distance between two probability distributions is to fit parametric density functions based on the available samples and calculate standard probability distance measures. However, for the sake of simplicity, we compute F-ratio between the distributions. Given two distributions $\mathcal{G}_0$ and $\mathcal{G}_1$, whose mean ($\mu_{0}$ and $\mu_{1}$, respectively) and standard deviation ($\sigma_{0}$ and $\sigma_{1}$, respectively) values are known, the F-ratio can be computed as:
% \vspace{-0.45cm}
\begin{equation}
     \mathbb{Q}(\mathcal{G}_0||\mathcal{G}_1) = \text{F-ratio}(\mathcal{G}_0,\mathcal{G}_1) = \frac{(\mu_{1} -\mu_{0} )^{2}}{\sigma_{1} ^{2}+\sigma_{0} ^{2}} \label{equation:Fratio}
\end{equation}
\noindent Since true mean and standard deviation values are unavailable, we estimate them from the available similarity scores. It is also possible to use other distance measures such as the Earth Mover's Distance (EMD) \cite{rubner2000earth}, which is defined as:
\begin{equation}
     \mathbb{Q}(\mathcal{G}_0||\mathcal{G}_1) = \text{EMD}(\mathcal{G}_0,\mathcal{G}_1) = \inf_{\nu \in \Gamma(\mathcal{G}_0,\mathcal{G}_1)} \int_{U \times V} |u-v| \, d\nu(u,v) \label{equation:EMD}
\end{equation}
\noindent where $\Gamma$ is the set of all joint distributions $\nu$ whose marginals are $\mathcal{G}_0$ and $\mathcal{G}_1$. 
%The integral computes the distance for this joint distribution $\nu$, integrating over all pairs of points $(u, v)$ from the respective distributions, weighted by the distance $|u-v|$ between them.
\noindent Finally, the normalization function $\mathcal{H}$ is selected as follows. Following the earlier discussion, in the worst-case scenario, the intra-class similarity distributions of the synthetic dataset would be close to that of the similarity distribution between images that are minor transformations of each other. Let $h$ denote a minor random transformation that can be applied to a real input image $\mathbf{x}_a^{r}$ to obtain a transformed image $\mathbf{x}_{a^{'}}^{r} = h(\mathbf{x}_a^{r})$. Let $S_{aa^{'}}^{r} = \mathcal{S}(\mathcal{F}(\mathbf{x}_a^{r}),\mathcal{F}(\mathbf{x}_{a^{'}}^{r}))$ be the similarity between a real input image and its transformed version. Let $\mathcal{G}_{trans}^{r}$ be the probability distribution of similarity score between a real image and its transformed counterpart. Finally, let $d_{max} = \mathbb{Q}(\mathcal{G}_{intra}^s||\mathcal{G}_{trans}^r)$. When $d_{intra}$ or $d_{inter}$ is closer to $d_{max}$, it indicates poor diversity. Therefore, we can employ the following exponential normalization function.

\begin{equation}
     \mathcal{H}_{\alpha}(d) = \exp\left(\ln{(\alpha)}\frac{d}{d_{max}}\right) \label{equation:Fratio}
\end{equation}

\noindent The above normalization function ensures that when $d \approx d_{max}$, $\mathcal{H}_{\alpha}(d) \approx \alpha$ and $\mathcal{H}_{\alpha}(d) \rightarrow 0$ when $d \rightarrow 0$. Here, $\alpha$ is usually set to a small value, say $10^{-4}$.

\noindent We use the samples from the given datasets $\mathcal{D}^s$ and $\mathcal{D}^r$ to empirically estimate the required distributions. For example, in the intra-class scenario, we select $n$ samples from a class and compute the similarities between all possible $n(n-1)/2$ pairs to obtain $\mathcal{G}_{intra}^{*}$. Since there are $M$ classes in the dataset, the number of possible similarity scores for the inter-class distribution will be ${n^2}*(M^2-M)$. Finally, to estimate $\mathcal{G}_{trans}^r$, we obtain a total of $nk$ similarity values for each class, where $k$ is the number of random transformations applied per image.
% \begin{figure}[h]
%     \centering
%     \includegraphics[width=0.7\textwidth]{Figures/examples.png}
%     \caption{\textbf{Examples of real and generated images} from the MIMIC-CXR and ImageNet datasets. For CXR images, a column consists of images from true data and generated data corresponding to the same underlying pathology. Likewise, for ImageNet, a column represents the same class label.}
%     \label{fig:examples}
% \end{figure}
% \vspace{-0.68cm}
\section{Experimental Results}
\noindent \textbf{Datasets and Generators}\label{sec:exp_prompts}: We use the MIMIC-CXR dataset, comprising 377,110 CXRs and associated reports, selecting representative samples from subsets p11, p12, and p13. The `impression' sections of the reports were analyzed with the CheXpert labeler to generate 14 diagnostic labels. We generated synthetic CXRs using the RoentGen \cite{chambon2022roentgen} with prompts crafted from CheXpert labels. Additionally, we matched 14 ImageNet classes with MIMIC-CXR classes for a broader evaluation, generated using UniDiffuser \cite{bao2023one} (Figure \ref{fig:examples}). Our experiments included three distinct prompt types for each dataset to investigate their impact on synthetic image quality and relevance. For MIMIC-CXR, $P_1=$ `${\texttt{CLS}}$', $P_2$ $=$ {\textit{`An image of a chest x-ray showing} ${\texttt{CLS}}$'}, and $P_3$ $=$ {\textit{`A realistic image of a chest x-ray showing} ${\texttt{CLS}}$'}, where \texttt{CLS} is the class name. For ImageNet, $P_1=$ `${\texttt{CLS}}$', $P_2$ $=$ {\textit{`An image of a} ${\texttt{CLS}}$'}, and $P_3$ $=$ {\textit{`A realistic image of a} ${\texttt{CLS}}$'}.\\

\noindent \textbf{Feature Extractor and Similarity Function}: We employed a ResNet50 backbone pre-trained on CXR using self-supervised contrastive learning for computing pairwise embeddings \cite{cho2023chess}. For ImageNet, a pre-trained ResNet50 was utilized to leverage its strong representation capabilities. Since our aim was to obtain representative embeddings, not to train or test the model, the potential bias concern is mitigated. The embeddings were used solely for similarity evaluation. Classifiers trained on synthetic ImageNet samples showed a significant drop in accuracy when tested on real data. Notably, there was a fourfold drop in accuracy when classifiers trained on synthetic CXRs were tested on real CXRs. This highlights the importance of evaluating the diversity of synthetic datasets. Our \texttt{SDICE} index correlates with these performance declines, offering valuable insights into the diversity of synthetic samples for downstream tasks.
\begin{figure}[t]
    \centering
    \begin{minipage}[b]{0.35\linewidth}  % Adjust width to fit page
        \centering
        \includegraphics[width=\linewidth]{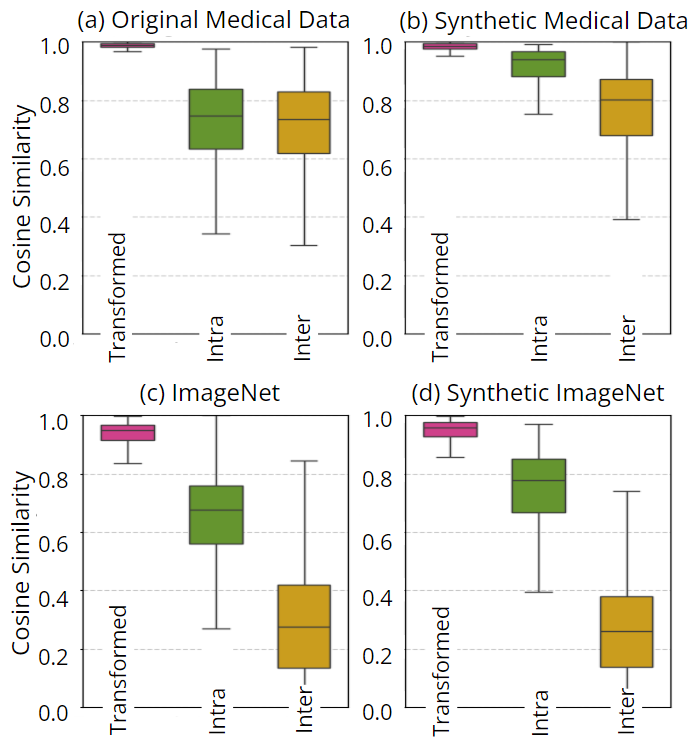}
        \vspace{3.9pt}
        \caption{Qualitative analysis of distribution change across cases}
        \label{fig:casewise_box}
    \end{minipage}
    \hspace{0.03\linewidth}  % Adjust the spacing between figures
    \begin{minipage}[b]{0.55\linewidth}  % Adjust width to fit page
        \centering
        \includegraphics[width=\linewidth]{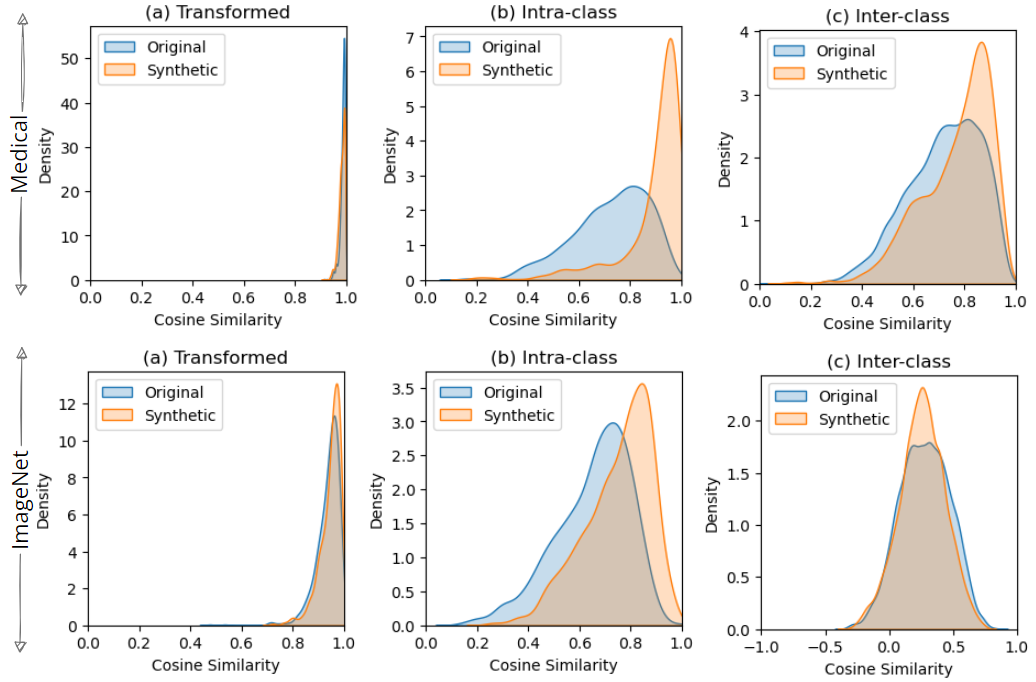}
        \vspace{3.9pt}
        \caption{Distribution variation and overlap between real and synthetic samples}
        \label{fig:kde}
    \end{minipage}
\end{figure}

% \begin{figure}[t]
%     \centering
%     \begin{subfigure}{0.1\textwidth}
%         \includegraphics[width=\textwidth]{Figures/casesise_box_new.png}
%         \caption{Qualitative analysis of distribution change across cases}
%         \label{fig:casewise_box}
%     \end{subfigure}
%     \hfill
%     \begin{subfigure}{0.1\textwidth}
%         \includegraphics[width=\textwidth]{Figures/KDE2.png}
%         \caption{Distribution variation between real and synthetic samples}
%         \label{fig:kde}
%     \end{subfigure}
%     \caption{\textbf{Similarity distributions} for both MIMIC-CXR and ImageNet datasets. (a) Box-plots showing the spread of the distributions across different cases, and (b) Density plots showing the distribution overlap between real and synthetic samples.}
%     % \vspace{-0.25cm}
% \end{figure}

\renewcommand{\arraystretch}{1.5}
\begin{table}[t]
\caption{\texttt{SDICE} index using F-ratio and EMD for intra (\(\gamma_{intra}\)) and inter (\(\gamma_{inter}\)) cases, along with the influence of sample size (a) and prompt type (b) on the \texttt{SDICE} index. Additional results on FairFace dataset \cite{karkkainenfairface} are provided in the supplementary material.}
\vspace{8.9pt}
\label{tab:combined}
\centering
\fontsize{7}{7}\selectfont
\resizebox{\textwidth}{!}{%
\begin{tabular}{l|l|cc|ccc|ccc} 
\hline
\rowcolor[HTML]{F3F2F2} 
\textbf{Dataset} & \textbf{  $\gamma$} & \multicolumn{2}{c|}{\textbf{\texttt{SDICE} index (\(\gamma\))}} & \multicolumn{3}{c|}{\textbf{Sample size}} & \multicolumn{3}{c}{\textbf{Prompt type}} \\ \cline{3-10}
\rowcolor[HTML]{F3F2F2} 
& & \textbf{F-ratio} & \textbf{EMD} & \(n\) & \(2n\) & \(4n\) & $P_1$ & $P_2$ & $P_3$ \\ \hline

\rowcolor[HTML]{FFFFFF} 
MIMIC \cite{johnson2019mimic}& \(\gamma_{intra}\) & 0.11 & 0.01 & 0.26 & 0.36 & 0.47 & 0.63 & 0.36 & 0.37 \\
% \rowcolor[HTML]{F3F2F2} 
& \(\gamma_{inter}\) & 0.84 & 0.44 & 0.74 & 0.84 & 0.99 & 0.91 & 0.84 & 0.89 \\ \hline

\rowcolor[HTML]{FFFFFF} 
ImageNet \cite{deng2009imagenet} & \(\gamma_{intra}\) & 0.47 & 0.26 & 0.37 & 0.47 & 0.56 & 0.47 & 0.83 & 0.89 \\
% \rowcolor[HTML]{F3F2F2} 
& \(\gamma_{inter}\) & 0.98 & 0.74 & 0.98 & 0.99 & 0.99 & 0.80 & 0.95 & 0.98 \\ \hline
\end{tabular}}
% \vspace{-0.25cm}
\end{table}

% \begin{figure*}
%     \centering
%     \includegraphics[width=0.80\textwidth]{Figures/Fisher_hist.png}
%     \includegraphics[width=0.15\textwidth]{Figures/abbreviations.png}
    
%     \caption{\textbf{Heatmap illustrating the difference between real and synthetic data distributions} for MIMIC-CXR and ImageNet, by cases and classes. It quantifies how closely synthetic data replicates the real, with varying degrees of diversity across both datasets.}
%     \label{fig:fisher}
%     % \vspace{-0.5cm}
% \end{figure*}

\begin{figure*}[t]
    \centering
    \includegraphics[width=1\textwidth]{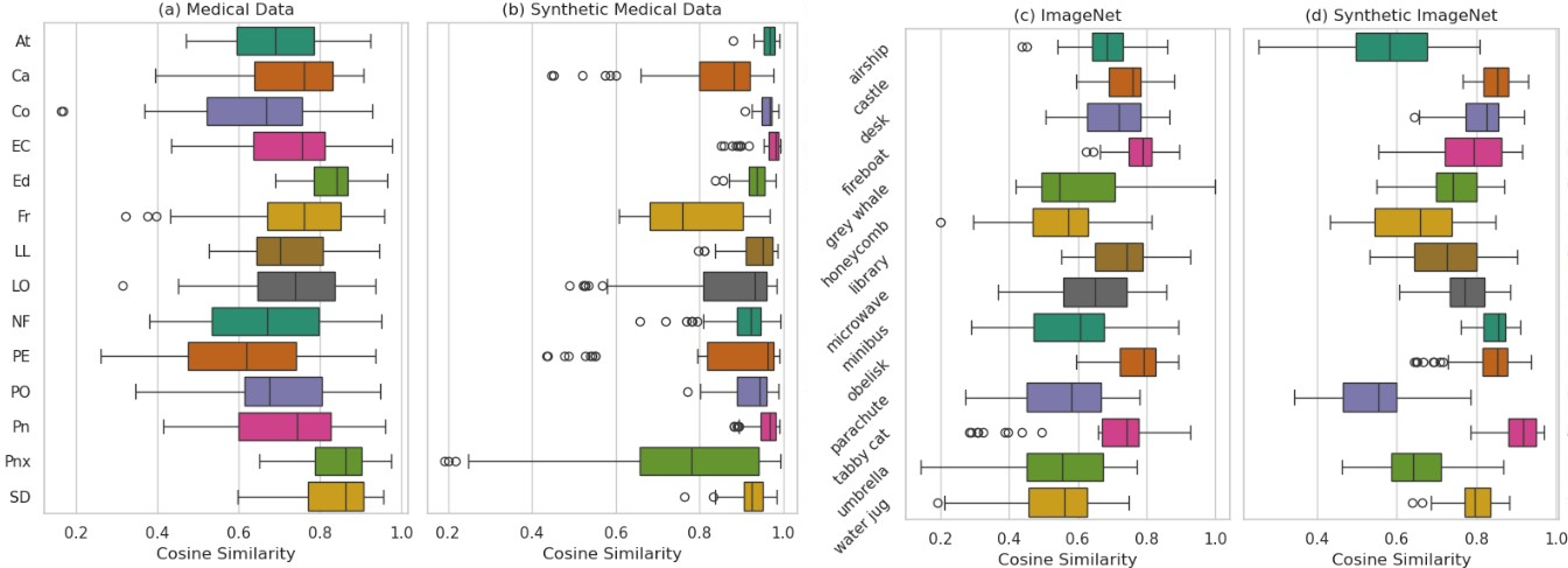}
    \vspace{0.0001pt}
    \caption{\textbf{Qualitative analysis of distribution differences between the real and synthetic samples in terms of individual classes.} (a) and (b) depict the class-wise distributions for the MIMIC-CXR dataset, while (c) and (d) illustrate the same for the 14 classes of ImageNet.}
    \label{fig:classwise_box}
\end{figure*}
\label{sec:Experiments}
% \vspace{-0.8cm}
\subsection{Diversity evaluation}
% We examined variations in cosine distribution for real and synthetic data across three types: transformation ($\mathcal{G}_{trans}^{*}$), intra-class ($\mathcal{G}_{intra}^{}$), and inter-class ($\mathcal{G}_{inter}^{}$).
% \textbf{Variations within Types.}
Firstly, we observe that $\mathcal{G}_{trans}^{r}$ consistently hovers close to 1.0 (see Figure \ref{fig:casewise_box}), which is expected because the feature extractor and similarity computation are designed to ignore differences between an image and its transformed version. However, intra-class variations depict a greater range in $\mathcal{D}^r$ than in $\mathcal{D}^s$ for CXRs, whereas inter-class variations present comparable extents across both distributions, as shown in Figure \ref{fig:kde}.
To quantitatively assess these observations, we computed the \texttt{SDICE} index using both F-ratio and EMD for each case under study. As detailed in Table \ref{tab:combined}, the \(\gamma_{intra}\) for CXRs is significantly lower, indicating a lack of intra-class diversity compared to those of ImageNet. An analysis using different sample sizes and prompts is also presented in Table \ref{tab:combined} and discussed in detail in section \ref{sec:ablation}.

% Figure \ref{fig:fisher} shows a heatmap illustrating the distance between real and synthetic data distributions, as measured by the F-ratio, across different cases and classes. 
Further investigation of \(\gamma_{intra}\) was done by measuring the diversity within individual classes of both datasets as shown in Figure \ref{fig:classwise_box} and detailed in Table \ref{tab:diversity_values}. We observe that several classes in the MIMIC-CXR synthetic dataset do not have the same range of diversity as its real counterpart. We observe poor diversity in classes with niche domain-specific names (such as `Atelectasis' and `Enlarged Cardiomegaly') as opposed to more general ones (`Pneumonia' and `Fracture'). We hypothesize that the generative model \cite{chambon2022roentgen} possibly fails to capture the true variations within the esoteric classes due to limited training. 
\begin{table}[t]
    \centering
    \begin{minipage}[c]{0.48\textwidth}  % Adjust the width of the minipage for the table
        \centering
        {\renewcommand{\arraystretch}{1.70}
        \scalebox{0.8}{  % Scale the table to 80% of its original size
        \begin{tabular}{lcclc}
        \rowcolor[HTML]{f3f2f2}
        \multicolumn{2}{c}{\textbf{MIMIC-CXR}} & \phantom{abc} & \multicolumn{2}{c}{\textbf{ImageNet}} \\
        \cmidrule{1-2} \cmidrule{4-5}
        Class & $\gamma_{intra}$ && Class & $\gamma_{intra}$ \\
        \midrule
        \cellcolor{pink}\textbf{At} & \cellcolor{pink}\textbf{1.56e-08} && airship & 3.53e-01 \\
        Ca & 4.29e-01 && castle & 1.19e-03 \\
        Co & 1.68e-05 && desk & 8.07e-02 \\
        EC & 1.43e-04 && \cellcolor{lightgreen}\textbf{fireboat} & \cellcolor{lightgreen}\textbf{9.97e-01} \\
        Ed & 1.74e-03 && grey whale & 9.01e-02 \\
        \cellcolor{lightgreen}\textbf{Fr} & \cellcolor{lightgreen}\textbf{7.93e-01} && honeycomb & 3.05e-01 \\
        LL & 3.25e-05 && library & 9.94e-01 \\
        LO & 3.91e-01 && microwave & 1.06e-01 \\
        NF & 3.38e-03 && \cellcolor{pink}\textbf{minibus} & \cellcolor{pink}\textbf{5.52e-05} \\
        PE & 5.24e-02 && obelisk & 5.32e-01 \\
        PO & 9.95e-04 && parachute & 9.54e-01 \\
        Pn & 3.14e-04 && tabby cat & 4.53e-03 \\
        Pnx & 5.56e-01 && umbrella & 3.21e-01 \\
        SD & 1.10e-01 && water jug & 1.36e-04 \\
        \bottomrule
        \end{tabular}}
        }  % End of scalebox
        \vspace{8.0pt}
        \caption{$\gamma_{intra}$ values where `Atelectasis' in \textbf{MIMIC-CXR} exhibits the least diversity, while `Fracture' demonstrates the highest diversity. In \textbf{ImageNet}, `minibus' class has the least diversity, and `fireboat' stands out as the most diverse class.}
        \label{tab:diversity_values}
    \end{minipage}%
    \hfill
    \begin{minipage}[c]{0.46\textwidth}  % Adjust the width of the minipage for the figures
        \centering
        \includegraphics[width=\linewidth]{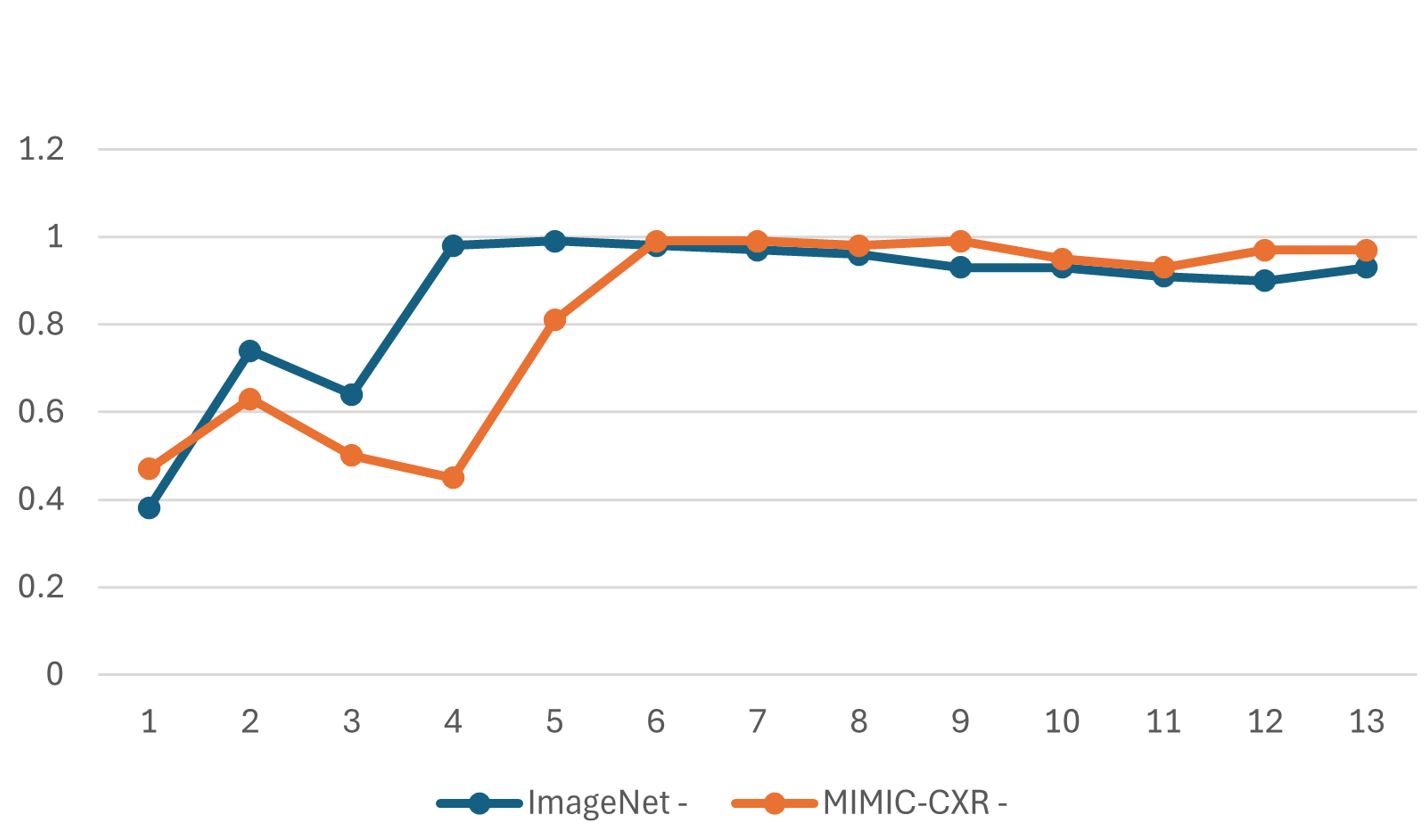}
        \vspace{0.01pt}
        \captionof{figure}{\textbf{Progression of \(\gamma_{inter}\) with increasing class inclusion.} This illustrates the increase in diversity within MIMIC-CXR and ImageNet as more classes are added, leveling off to indicate a maximum inter-class diversity threshold.}
        \label{fig:inter_prog}
        \vspace{0.2cm} % Adjust vertical spacing between figures
        \includegraphics[width=\linewidth]{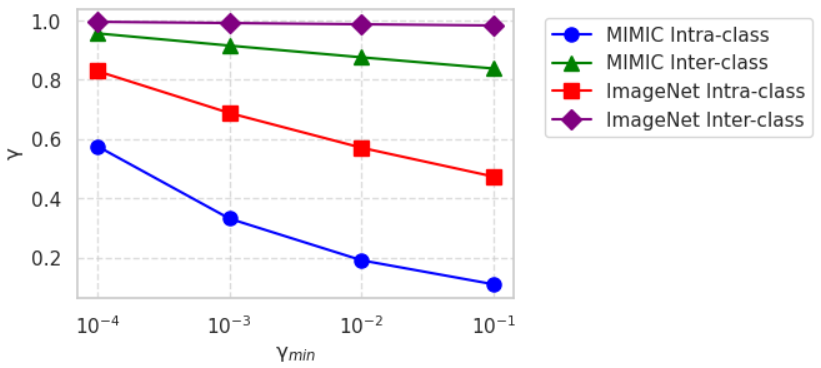}
        \vspace{2.9pt}
        \captionof{figure}{\textbf{\texttt{SDICE} Index Variation with $\gamma_{min}$ in MIMIC-CXR and ImageNet.} This displays a marked decrease in MIMIC-CXR's intra-class diversity with increasing $\gamma_{min}$, in contrast to ImageNet's consistent inter-class diversity.}
        \label{fig:gamma_min}
    \end{minipage}
\end{table}
Inter-class variation in the generated data is similar to that of the reference data for both datasets. To confirm the observed trend in \(\gamma_{inter}\) across datasets, we divided the dataset into `$q$' subsets where `$q$' is the number of classes in the dataset. Each subset gradually included more classes, starting with only the first class in the initial subset and progressing to the last set with all available classes. Our findings (Figure \ref{fig:inter_prog}) reveal a gradual increase in \(\gamma_{inter}\), indicating growing diversity. However, this upward trend reaches a saturation point suggesting that inter-class diversity does not increase beyond a certain threshold of class inclusion.
% \begin{figure}
%     \centering
%     \begin{minipage}{.47\textwidth}
%         \centering
%         \includegraphics[width=\linewidth]{Figures/inter_prog.JPG}
%         \caption{\textbf{Progression of \(\gamma_{inter}\) with increasing class inclusion.} This illustrates the increase in diversity within MIMIC-CXR and ImageNet as more classes are added, leveling off to indicate a maximum inter-class diversity threshold.}
%         \label{fig:inter_prog}
%     \end{minipage}\hfill
%     \begin{minipage}{.5\textwidth}
%         \centering
%         \includegraphics[width=\linewidth]{Figures/gamma_min_vs_gamma.png}
%         \caption{\textbf{\texttt{SDICE} Index Variation with $\gamma_{min}$ in MIMIC-CXR and ImageNet.} This displays a marked decrease in MIMIC-CXR’s intra-class diversity with increasing $\gamma_{min}$, in contrast to ImageNet's consistent inter-class diversity.}
%         \label{fig:gamma_min}
%     \end{minipage}
%     % \vspace{-0.5cm}
% \end{figure}
\subsection{Comparison with SSIM and FID}
% In generative model evaluation, fidelity and diversity are central metrics, commonly assessed via FID and SSIM scores. However, the FID score, derived using an Inception V3 model pre-trained on ImageNet, may falter in domain-specific settings such as CXRs, due to its potential inability to accurately represent domain-relevant features. This assertion is supported by our calculations of the mean FID scores for both inter and intra similarity distributions, yielding values of 0.0082 and 0.0099, respectively. Such results do not offer definitive conclusions, unlike those derived from our \texttt{SDICE} index which utilizes domain-specific contrastive encoder, which aligns well with our analytical expectations. The FID-derived intra and inter cases are indistinguishably close, lacking clear differentiation. 
% Similarly, the mean SSIM values for intra and inter cases, specifically 0.68 and 0.60, do not present a clear distinction, highlighting the limitations of these metrics. 
% In the evaluation of generative models, traditional metrics like FID and SSIM scores are integral for assessing fidelity and diversity. However, these metrics may not fully capture domain-specific features, as seen with chest X-rays (CXR), where FID scores derived from ImageNet-trained models fall short. 
A further analysis was conducted to highlight the variation obtained by the \texttt{SDICE} index as compared to the SSIM and FID scores (Supp: Table \ref{tab:ssim_fid_scores}). Our analysis shows mean FID scores hover around 0.0082 and 0.0099 for intra and inter-class distributions, respectively. The FID score shows poor resolution as compared to the \texttt{SDICE} index, as the latter benefits from a domain-specific contrastive encoder. Similarly, SSIM values also fail to provide a clear separation between intra and inter-class diversity with mean SSIM scores of 0.68 and 0.60, respectively. The \texttt{SDICE} index effectively highlights the contrast in diversity across intra and inter-class categories, providing clear insights that FID and SSIM metrics may overlook. This highlights the benefits of \texttt{SDICE} index in domain-specific dataset analysis.
% In contrast, our \texttt{SDICE} index provides detailed and conclusive metrics, distinctly outlining the differences between intra and inter-class diversity, thereby validating insights from our prior analysis. This differentiation is not easily evident when relying solely on FID and SSIM scores, emphasizing the enhanced analytical utility of the \texttt{SDICE} index within our domain-specific context.
\subsection{Ablation Studies}
\label{sec:ablation}
\subsubsection{Impact of number of samples on \(\gamma_{intra}\)} Table \ref{tab:combined}(a) outlines how \(\gamma_{\text{intra}}\) values evolve as we increase sample sizes from \(n\) to \(2n\), and further to \(4n\). This progression reveals that \(\gamma_{\text{intra}}\), or the measure of diversity within classes, tends to rise with larger sample sets. Observed in both the MIMIC and ImageNet datasets, this trend suggests that expanding the dataset by introducing a wider variety of examples within each class enhances the overall diversity. The initial sample size was 350 for both MIMIC-CXR and ImageNet datasets, meaning 25 images per class. We found that a balanced sample size yielded better results in terms of diversity assessment compared to imbalanced samples.

\subsubsection{Impact of different prompts on \(\gamma_{intra}\)} Table \ref{tab:combined}(b) shows that the complexity of prompts affects the diversity of the generated images. In the case of CXR images, less detailed prompts, such as $P_1$, appear to encourage a wider diversity, perhaps due to the generative model having broader interpretative freedom. For ImageNet, descriptive prompts such as $P_3$ lead to more diverse outputs, which implies that the detailed nature of these prompts provides useful guidance to the model, enabling it to capture the extensive variability inherent across ImageNet's classes. This suggests that the level of detail in prompts should be carefully considered to match the desired diversity of the dataset being synthesized.
\subsection{Parameter Sensitivity Analysis}
We investigate how variations in the parameter $\gamma_{min}$ affect the \texttt{SDICE} index ($\gamma$) in both the MIMIC-CXR and ImageNet datasets. Figure \ref{fig:gamma_min} illustrates the sensitivity of the \texttt{SDICE} index to changes in $\gamma_{min}$. The figure reveals a clear downward trend in intra-class scenarios for both ImageNet and MIMIC-CXR, indicating lower intra-class diversity compared to inter-class diversity. Notably, MIMIC-CXR consistently exhibits significantly lower intra-class diversity values compared to ImageNet across different $\gamma_{min}$ values. In the inter-class scenario, MIMIC-CXR shows a more pronounced downward trend compared to the stable and consistently diverse inter-class diversity observed in ImageNet (around $\gamma$ = 1.0) across various $\gamma_{min}$ values. This emphasizes how the \texttt{SDICE} index is sensitive to parameter changes, revealing distinct diversity characteristics within datasets.
% \subsection{Qualitative Diversity Assessment}
% We conducted a visual assessment of images from the MIMIC-CXR and ImageNet datasets, along with their corresponding synthetic counterparts generated using the Roentgen and Stable Diffusion models, respectively. This analysis focused on the synthetic images of the most and least diverse classes identified using our proposed \texttt{SDICE} index.
% As shown in Figure \ref{fig:classwise_box} `Atelectasis' is identified as the least diverse class, while `Fracture' is the most diverse class in CXR generation. For natural image generation, `minibus' is the class with the least diversity, and `fireboat' is the most diverse class according to the \texttt{SDICE} index.
% Figure \ref{fig:supp_1} further illustrates these findings: all synthetic images generated for the Atelectasis' class look very similar, confirming their lower diversity. Conversely, the synthetic images of the Fracture' class exhibit considerable diversity, accurately captured by the \texttt{SDICE} index. Similarly, synthetic `minibus' images show close similarity in color and shape, unlike their real counterparts, which justifies their low \texttt{SDICE} index.
% \vspace{-0.25cm}
\section{Conclusion}
% In this work, we introduced the \texttt{SDICE} index to evaluate the diversity captured by generative models, assessing variations at the intra- and inter-class levels. Our findings reveal that \texttt{SDICE} provides a more detailed perspective beyond traditional metrics like MS-SSIM and FID, highlighting a notable similarity and potential diversity deficit in synthetic Chest XRays compared to synthetic ImageNet samples. The study also underscores the impact of sample size and prompt descriptiveness on the diversity of the two datasets.
This work introduced the \texttt{SDICE} index for evaluating the diversity of synthetic medical image datasets. Leveraging the power of contrastive encoders, the \texttt{SDICE} index characterizes the similarity distributions observed in the reference and synthetic datasets and provides a normalized measure to assess and compare dataset variability. Our experiments on MIMIC-CXR and ImageNet confirm its efficacy, revealing particularly low diversity in synthetic CXRs, highlighting areas where generative models may need refinement. Moving forward, we will focus on reducing the computational complexity of the similarity score computation by exploring more efficient methods, such as approximate nearest neighbors. These improvements aim to enhance the scalability and practicality of our approach, further solidifying the \texttt{SDICE} index in the evaluation of synthetic medical image datasets.
% In this paper, we introduced \texttt{SDICE}, a new index designed to evaluate how well generative models capture diversity. Grounded in the concept of distribution variations, \texttt{SDICE} allows for a thorough assessment of diversity at both the macro and class-wise levels. Our experiments demonstrate that \texttt{SDICE} index offers a more detailed perspective compared to traditional metrics like MS-SSIM and FID. The results showed a significant degree of similarity among generated CXR images, indicating a potential lack of diversity in synthetic CXRs compared to ImageNet-generated samples. We also observed how the number of samples influences diversity, with an increasing trend for non-medical datasets but a diverse pattern for medical datasets. We also observed that more descriptive prompts enhance diversity for non-imaging datasets but reduce diversity in CXRs. 
\label{sec:Conclusion}
% \newpage
% \bibliography{main}

% \clearpage
% \bibliographystyle{unsrt}
\bibliography{egbib}

% \pagebreak
\newpage
\appendix
\noindent\section{\begin{huge} \textbf{Supplementary Material} \vspace{4mm} \end{huge}}

\begin{figure}[h]
    \centering
    \includegraphics[width=0.8\textwidth]{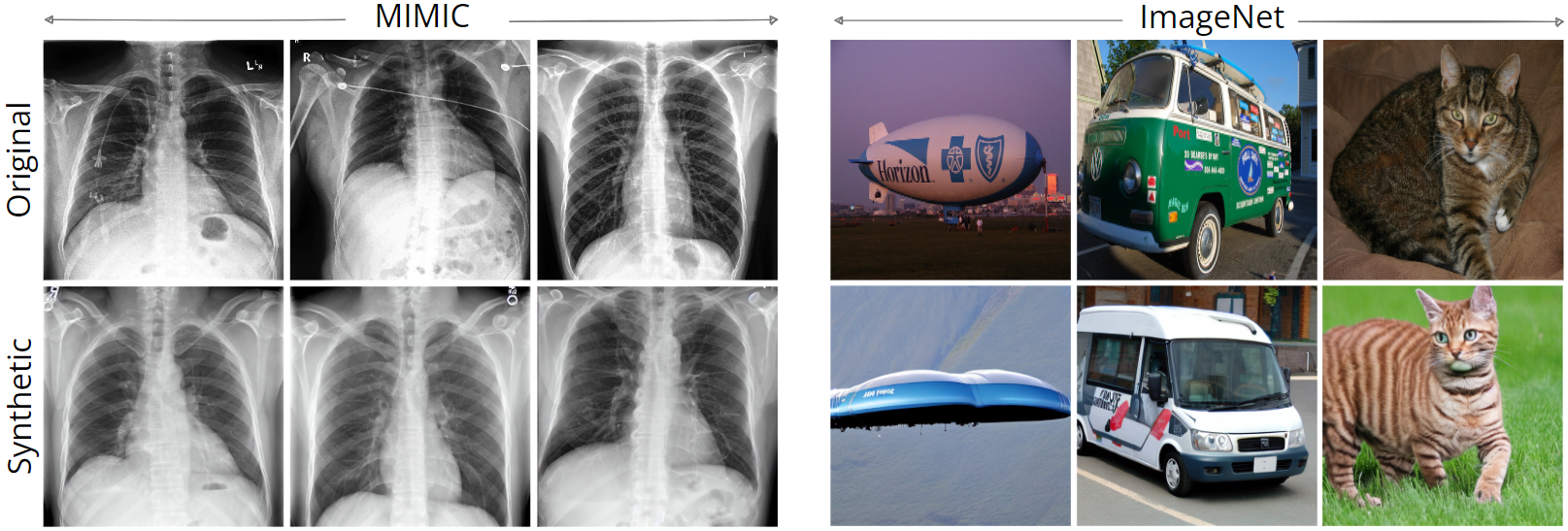}
    \vspace{9.9pt}
    \caption{\textbf{Examples of real and generated images} from the MIMIC-CXR and ImageNet datasets. For CXR images, a column consists of images from true data and generated data corresponding to the same underlying pathology. Likewise, for ImageNet, a column represents the same class label.}
    \label{fig:examples}
\end{figure}

\begin{figure*}[htp]
    \centering
    \includegraphics[width=0.83\textwidth]{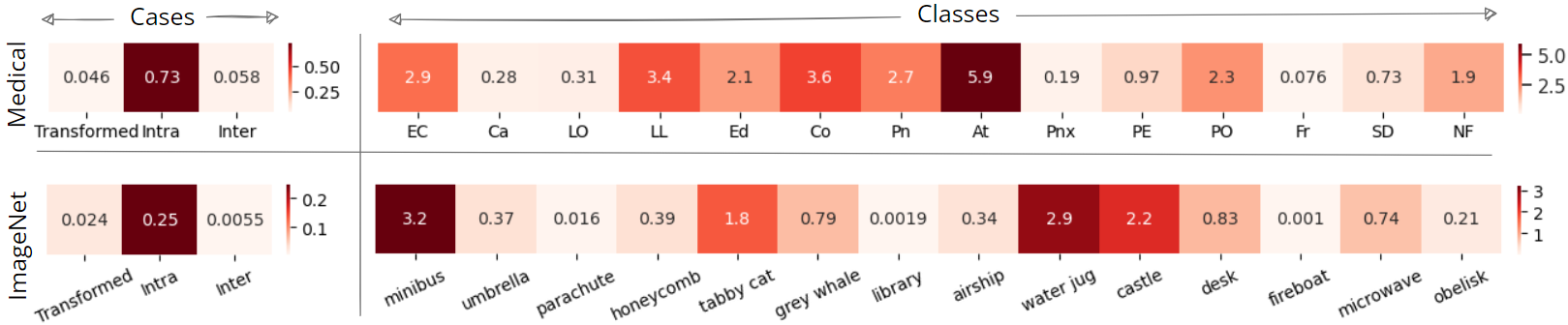}
    \includegraphics[width=0.15\textwidth]{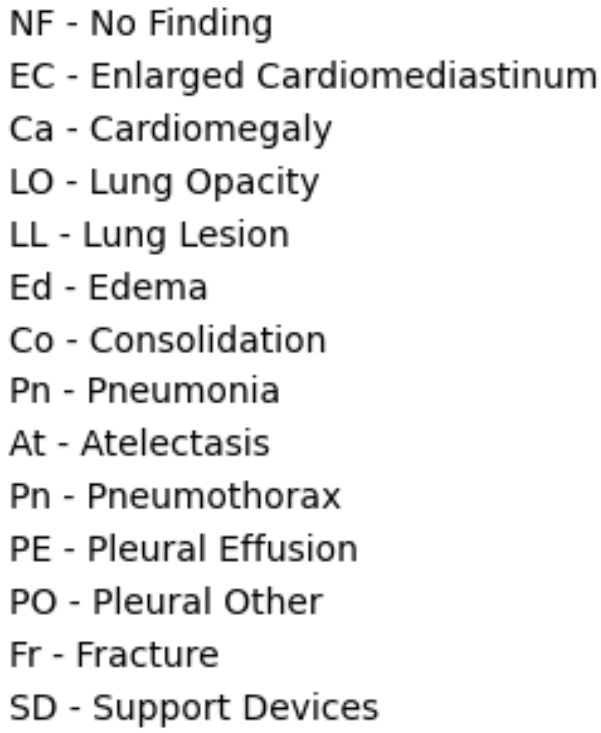}
    \vspace{9.9pt}
    \caption{\textbf{Heatmap illustrating the difference between real and synthetic data distributions} for MIMIC-CXR and ImageNet, by cases and classes. It quantifies how closely synthetic data replicates the real, with varying degrees of diversity across both datasets.}
    \label{fig:fisher}
    % \vspace{-0.5cm}
\end{figure*}

\begin{table}[htbp]
  \centering
  % \scriptsize % Reduce the font size of the table
  \setlength{\tabcolsep}{8pt} % Adjust the space between columns
  \renewcommand{\arraystretch}{1.0} % Adjust the height of each row
  \caption{Analysis of SSIM and FID Scores. This table shows SSIM and FID scores, highlighting their relative ineffectiveness in distinguishing between intra and inter-class diversity, as opposed to the \texttt{SDICE} index, which provides a clearer distinction of intra and inter-class diversity in both datasets.\\}
  \label{tab:ssim_fid_scores}
  \resizebox{0.45\textwidth}{!}{%
  \begin{tabular}{lccc}
    \toprule
    \rowcolor[HTML]{f3f2f2}
\textbf{Dataset} & \textbf{Case} & \textbf{Mean} & \textbf{Mean} \\
\rowcolor[HTML]{f3f2f2}
\textbf{} & \textbf{} & \textbf{SSIM} & \textbf{FID} \\
    \midrule
    MIMIC & Intra & 0.68 & 0.0082 \\
          & Inter & 0.60 & 0.0099 \\
    \addlinespace
    ImageNet & Intra & 0.30 & 0.17 \\
             & Inter & 0.06 & 0.20 \\
    \bottomrule
  \end{tabular}}
\end{table}

\subsection{Additional Empirical Studies on the FairFace Dataset}
This experiment validates the \texttt{SDICE} index as a straightforward yet effective way to measure diversity in synthetic datasets. It reliably gave higher diversity scores for datasets that accurately reflected the ethnic faces of their real-world counterparts. The highest \texttt{SDICE} score was observed in the mixed ethnicity ('Asian', 'Black' and 'White') dataset, showcasing the index's versatility. Such a tool is valuable for creating balanced datasets, which is crucial for building fair and unbiased AI systems. The \texttt{SDICE} index also helps identify where datasets might be lacking in diversity, providing a clear path for improvement and ensuring that AI models trained on these datasets perform well across diverse populations.
\begin{figure}[htp]
  \centering

  % Asian faces
  \begin{minipage}{0.49\textwidth}
    \centering
    \includegraphics[width=\linewidth]{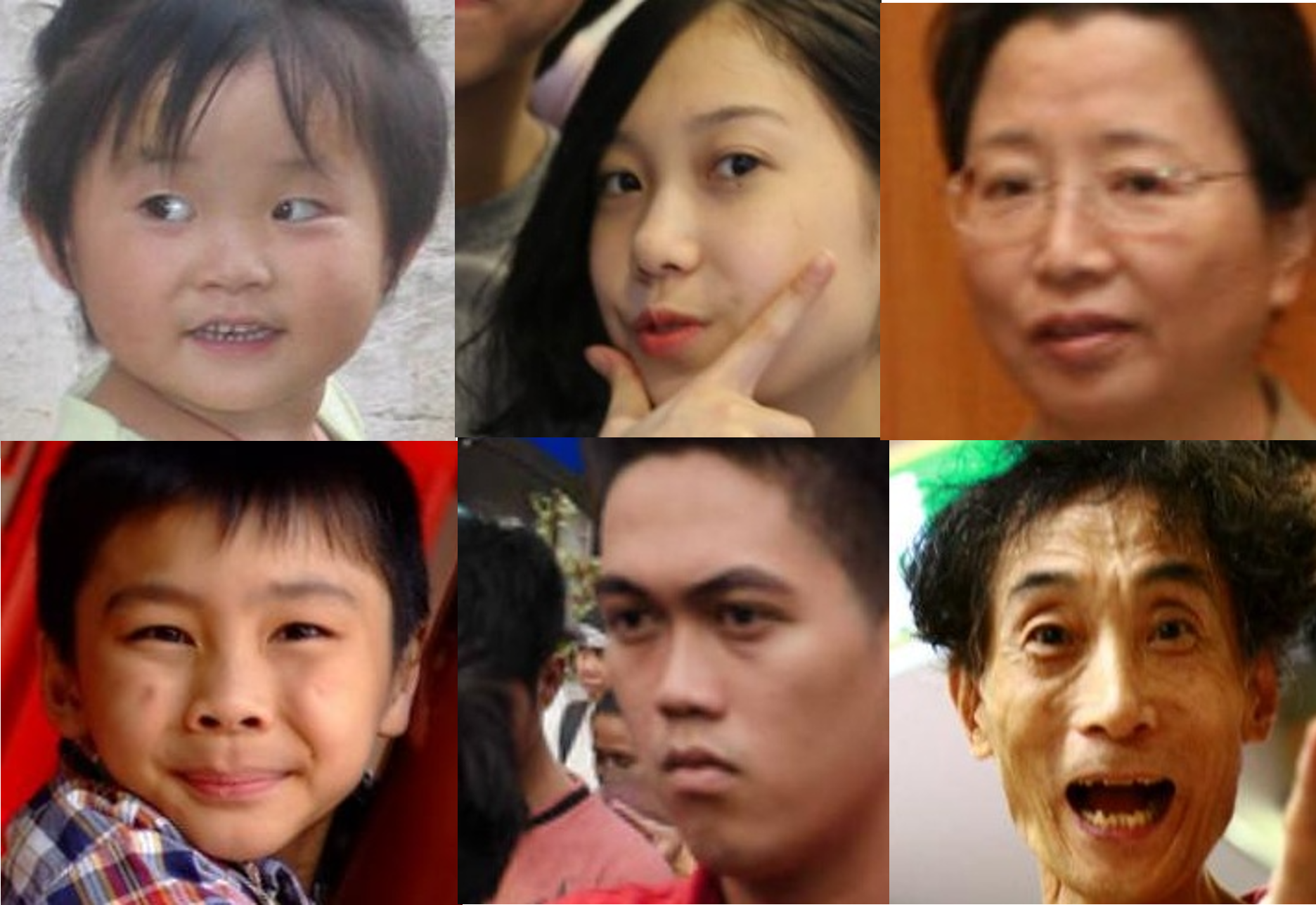}
    % \caption*{Asian}
    Asian
  \end{minipage}%
  \hfill
  \begin{minipage}{0.49\textwidth}
    \centering
    \includegraphics[width=\linewidth]{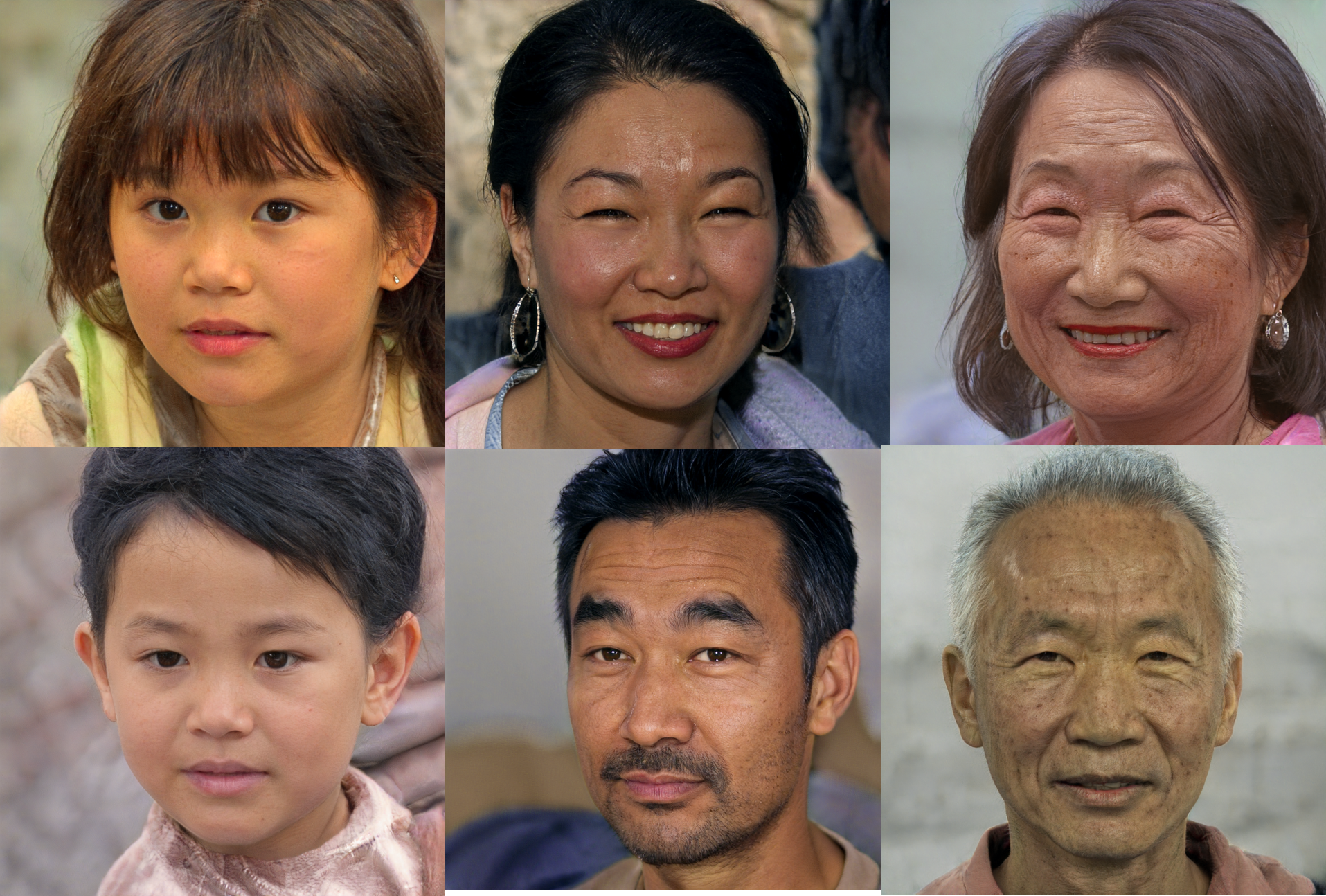}
    % \caption*{Corresponding Synthetic Dataset}
    Corresponding Synthetic Dataset
  \end{minipage}
  
  \vspace{0.3cm}  % Add vertical space between rows

  % White faces
  \begin{minipage}{0.49\textwidth}
    \centering
    \includegraphics[width=\linewidth]{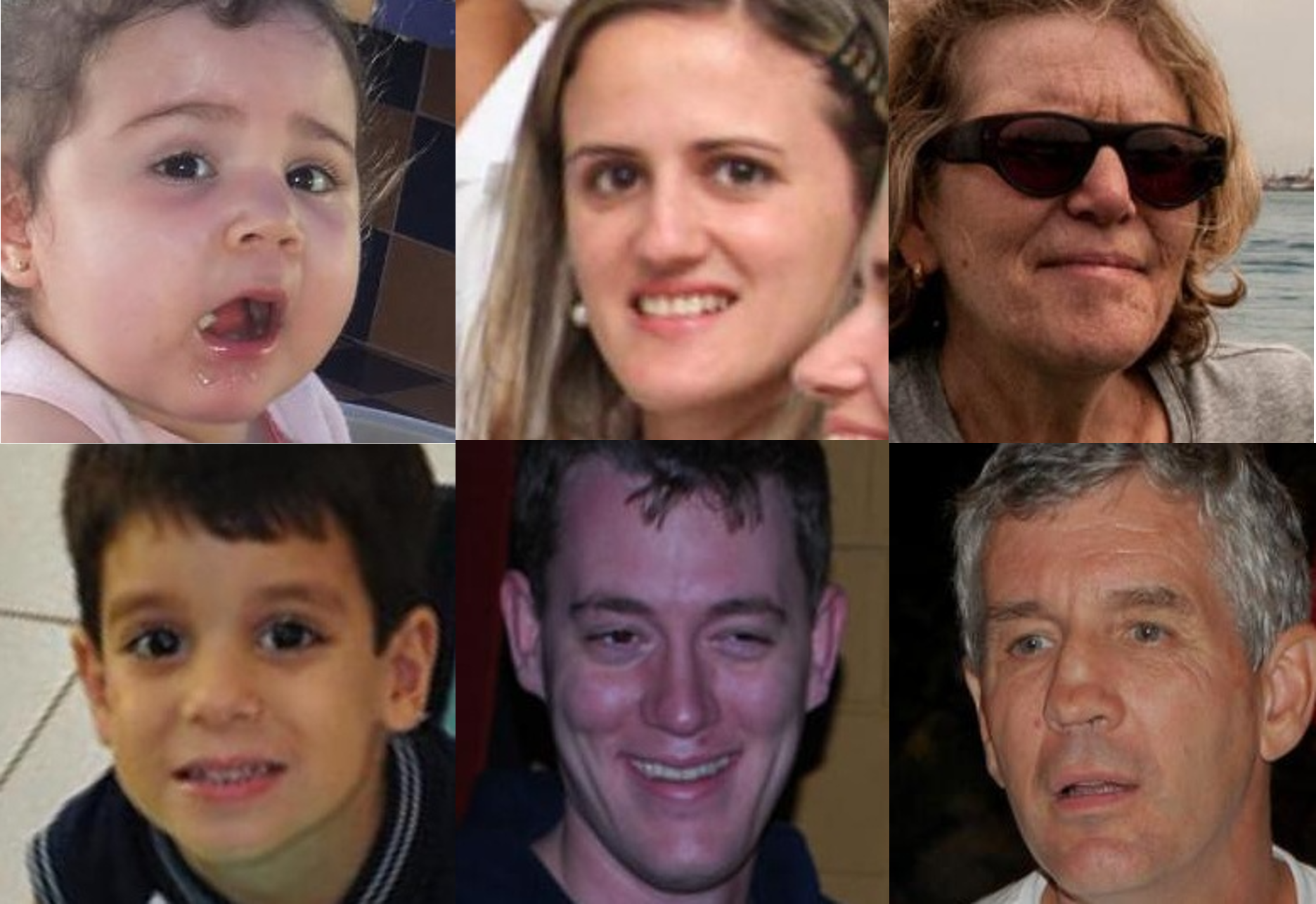}
    % \caption*{White}
    White
  \end{minipage}%
  \hfill
  \begin{minipage}{0.49\textwidth}
    \centering
    \includegraphics[width=\linewidth]{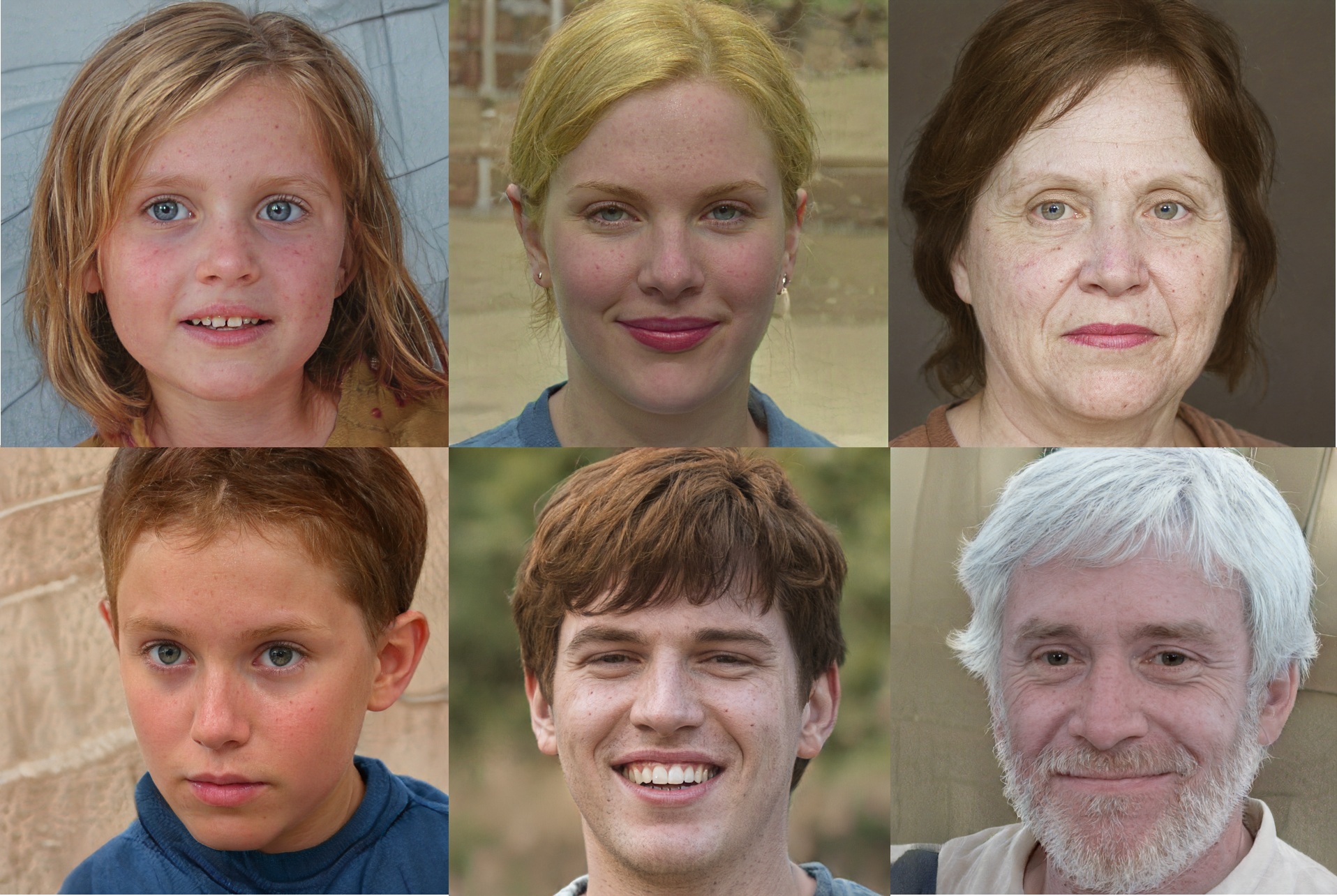}
    % \caption*{Corresponding Synthetic Dataset}
    Corresponding Synthetic Dataset
  \end{minipage}
  
  \vspace{0.3cm}  % Add vertical space between rows

  % Black faces
  \begin{minipage}{0.49\textwidth}
    \centering
    \includegraphics[width=\linewidth]{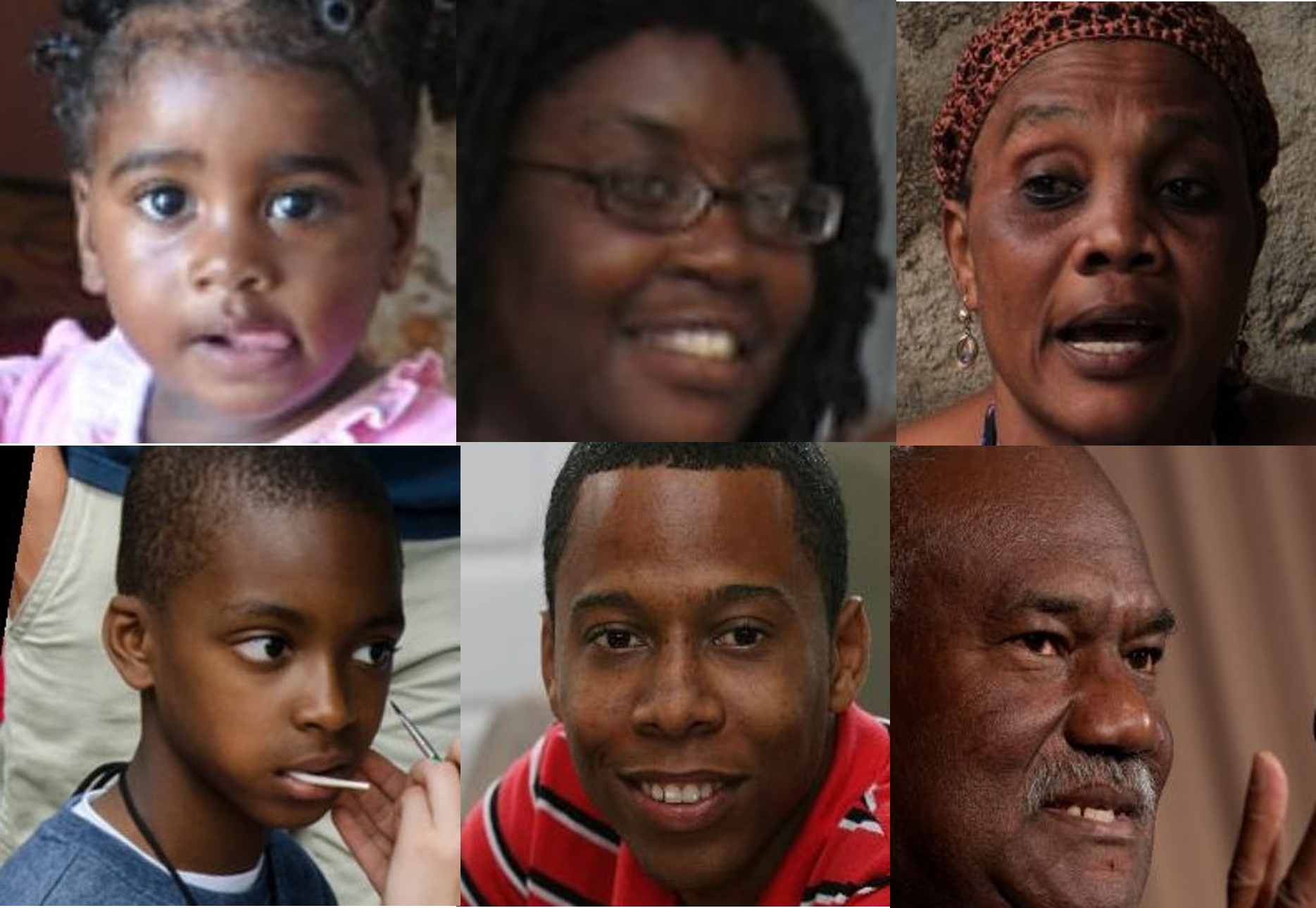}
    % \caption*{Black}
    Black
  \end{minipage}%
  \hfill
  \begin{minipage}{0.49\textwidth}
    \centering
    \includegraphics[width=\linewidth]{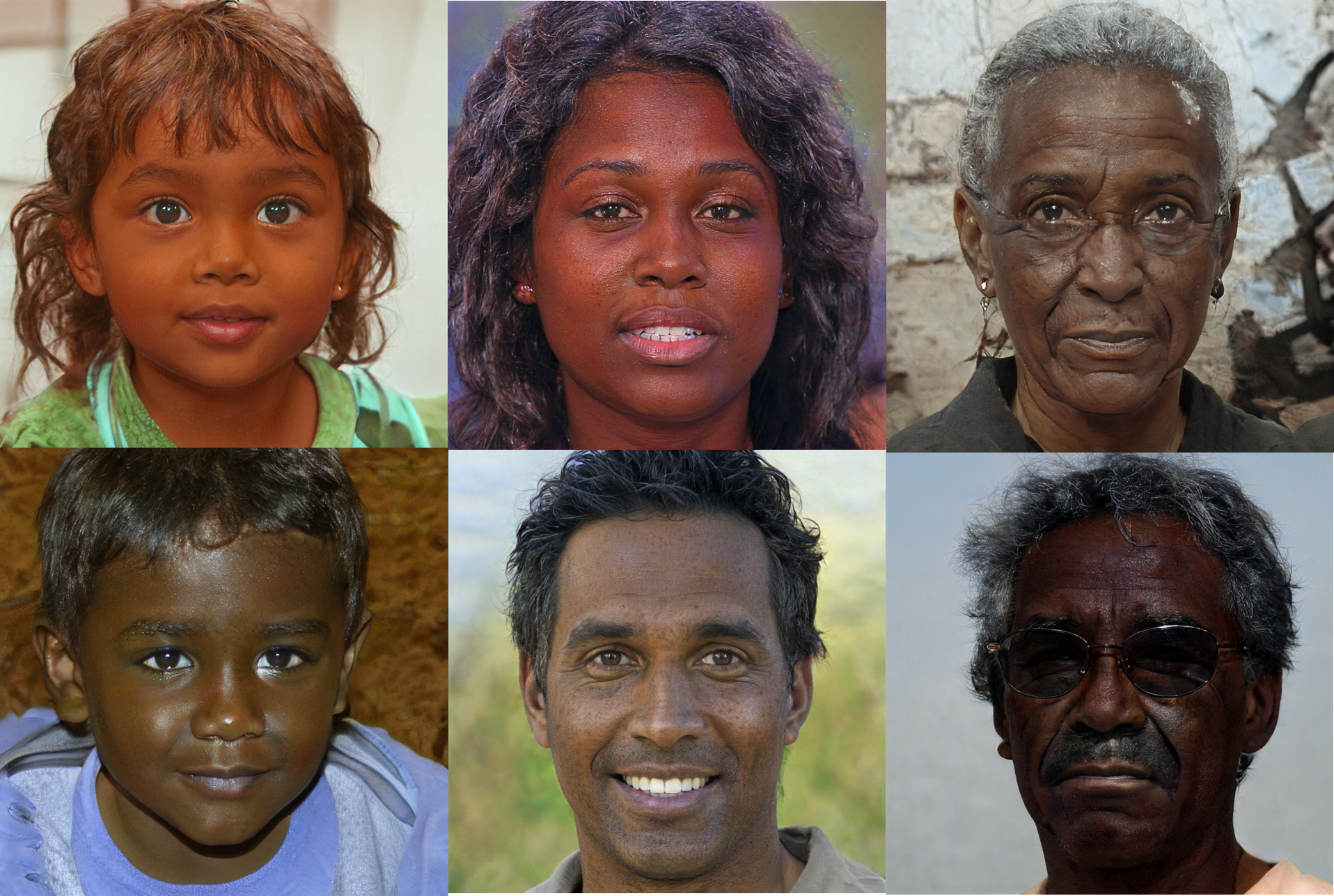}
    % \caption*{Corresponding Synthetic Dataset}
    Corresponding Synthetic Dataset\\
  \end{minipage}
  \vspace{0.8cm}
  \caption{Comparative Analysis of Face Images across Asian, Black, and White Races Featuring Diverse Age Groups and Genders in Original (FairFace) vs. Synthetic Datasets generated via StyleGAN2}
  \label{fig:races}
\end{figure}

\begin{table}[htbp]
  \centering
  % \scriptsize % Reduce the font size of the table
  \setlength{\tabcolsep}{8pt} % Reduce the space between columns
  \renewcommand{\arraystretch}{1.0} % Reduce the height of each row slightly
  \caption{Performance of \texttt{SDICE} index on the FairFace dataset, evaluating diversity in facial attributes across three distinct ethnicities: Asian, Black, and White. The table presents the calculated \texttt{SDICE} index values for synthetic images compared to their original counterparts, offering insights into the effectiveness in capturing diversity in the context of diverse facial characteristics.\\}
  \vspace{3.0pt}
  \resizebox{0.75\textwidth}{!}{%
  \begin{tabular}{lcccc}
    \toprule
    \rowcolor[HTML]{f3f2f2}
    & \multicolumn{4}{c}{\textbf{Reference Dataset}} \\
    % \cmidrule(lr){2-5}
    \rowcolor[HTML]{f3f2f2}
    \textbf{Synthetic} & Asian & Black & White & Overall (AWB) \\
    \midrule
    Asian & 0.994 & - & - & 0.866 \\
    Black & - & 0.994 &  - & 0.977 \\
    White & - & - & 0.999 & 0.731 \\
    Asian-Black & 0.963 & 0.964 & - & 0.727 \\
    Asian-White & 0.857 & - & 0.833 & 0.974 \\
    Black-White & - & 0.831 & 0.895 & 0.965 \\
    Asian-White-Black & 0.901 & 0.949 & 0.768 & \textbf{0.988} \\
    \bottomrule
  \end{tabular}}
\end{table}

%, aiming to identify the most and least diverse classes. Following our results from  we observe that, within the , . In the ImageNet dataset, the class 'minibus' displays the least diversity, in contrast to the class 'fireboat,' which exhibits the highest level of diversity.
%This involves a comparative analysis between original images and their synthetic counterparts. This qualitative assessment depicted in Figure \ref{fig:supp_1} allows us to visually assess the effectiveness of the synthetic generation models in capturing and representing the diversity inherent in different classes across both medical and natural image datasets.

\subsection{Impact of Feature Extractor}
\begin{figure}[htp]
    \centering
    \includegraphics[width=0.7\textwidth]{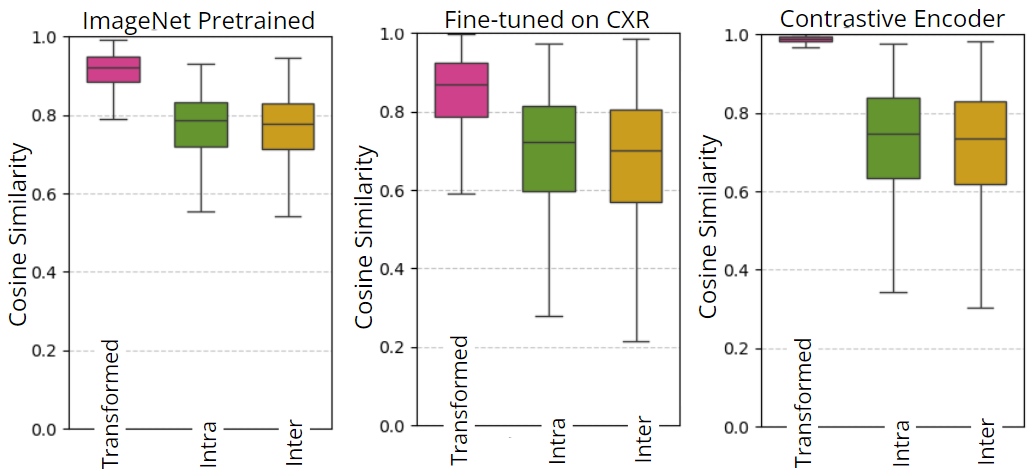}
    \vspace{8.9pt}
    \caption{Comparative performance of three feature extractors: ResNet-50 pre-trained on ImageNet, ResNet-50 pre-trained on ImageNet and fine-tuned on MIMIC-CXR, and ResNet-50 pre-trained on MIMIC-CXR using self-supervised contrastive learning. Contrastively learned encoder provides results that are closest to the expected behavior - similarity scores for different transformations of the same image are close to 1.}
    \label{fig:supp_encoder}
\end{figure}

One of the key factors impacting the proposed diversity assessment framework is the feature extractor $\mathcal{F}$. We examined the performance of three feature extractors having the same architecture (ResNet-50), but trained in different ways. We consider: (i) ResNet-50 model pre-trained on the ImageNet dataset, (ii) ResNet-50 model that is pre-trained on the ImageNet dataset and fine-tuned on MIMIC-CXR, and (iii) ResNet-50 model that is pre-trained using self-supervised contrastive learning on CXR images. In all the three cases, the cosine similarities are calculated for the transformed, intra-class, and inter-class scenarios, and the corresponding box plots are shown in Figure \ref{fig:supp_encoder}. \\

A closer analysis of Figure \ref{fig:supp_encoder} indicates that the contrastive learning encoder is the best choice for assessing diversity. This is because it gives similarity scores closest to 1 (compared to the other feature extractors) for the transformed case, which is the expected behavior. Notably, in the transformed case, the similarity scores are closest to 1 compared to the other encoders. This observation indicates that the contrastive learning encoder excels in preserving and understanding image representations, emphasizing its effectiveness in our diversity evaluation framework. 

\subsection{Impact of Distance Metric}
The proposed \texttt{SDICE} index employs the F-ratio as a primary distance measure between two probability distributions ($\mathcal{G}_0$ and $\mathcal{G}_1$) as defined in Eq. \ref{equation:Fratio}. To explore the robustness and sensitivity of the \texttt{SDICE} index, we also consider an alternative distance measure based on the Euclidean distance \cite{scipyemd}. Table \ref{tab:sdice_index} illustrates the impact of these distance metrics on the \texttt{SDICE} index $\gamma$ across different datasets.

For the MIMIC dataset, when evaluating intra-class diversity, the intra-class $\gamma$ values using the Cosine approach reveal significant differences between the F-ratio (0.11) and Euclidean (0.60) metrics. This disparity suggests that the choice of distance metric can heavily influence the perceived diversity within the same class, with F-ratio potentially capturing more subtle variations that Euclidean distance might overlook. The higher inter-class diversity indicated by the F-ratio further supports its sensitivity to variations between different classes compared to Euclidean, underscoring its utility in distinguishing between diverse groups.

Similarly, in the ImageNet dataset, the intra-class $\gamma$ values using the Cosine approach differ markedly between F-ratio (0.47) and Euclidean (0.03), again highlighting the significant impact that the choice of metric has on diversity assessment within the same class. When examining inter-class diversity, the F-ratio (0.98) suggests a much higher level of diversity between different classes than Euclidean (0.45). This finding indicates that while both metrics reveal similar trends, F-ratio is more effective in capturing the complexities of inter-class diversity, making it a more sensitive measure for evaluating how well the synthetic data represents the varied nature of the real-world datasets.
\begin{table}[htbp]
  \centering
  \caption{Comparison of \texttt{SDICE} index (\(\gamma\)) using F-ratio and Euclidean distance metrics for evaluating Intra-Class and Inter-Class Diversity in the MIMIC and ImageNet datasets. \\}
  % \vspace{5.0}
  \label{tab:sdice_index}
  \small  % Reduce the font size
  \setlength{\tabcolsep}{8pt}  % Reduce the column separation
  \renewcommand{\arraystretch}{1.0}  % Adjust the row spacing
  \resizebox{0.55\textwidth}{!}{%
  \begin{tabular}{lccc}
    \toprule
    \rowcolor[HTML]{f3f2f2}
    & & \multicolumn{2}{c}{\textbf{SDICE Index (\(\gamma\))}} \\
    % \cmidrule(lr){3-4}
    \rowcolor[HTML]{f3f2f2}
    Dataset & Case & F-ratio & Euclidean \\
    \midrule
    MIMIC & \(\gamma_{intra}\) & 0.11 & 0.60 \\
          & \(\gamma_{inter}\) & 0.83 & 0.86 \\
    \addlinespace
    ImageNet & \(\gamma_{intra}\) & 0.47 & 0.03 \\
             & \(\gamma_{inter}\) & 0.98 & 0.45 \\
    \bottomrule
  \end{tabular}}
\end{table}
\subsection{Impact of Guidance Scale}
\begin{figure*}
    \centering
    \includegraphics[width=0.9\textwidth]{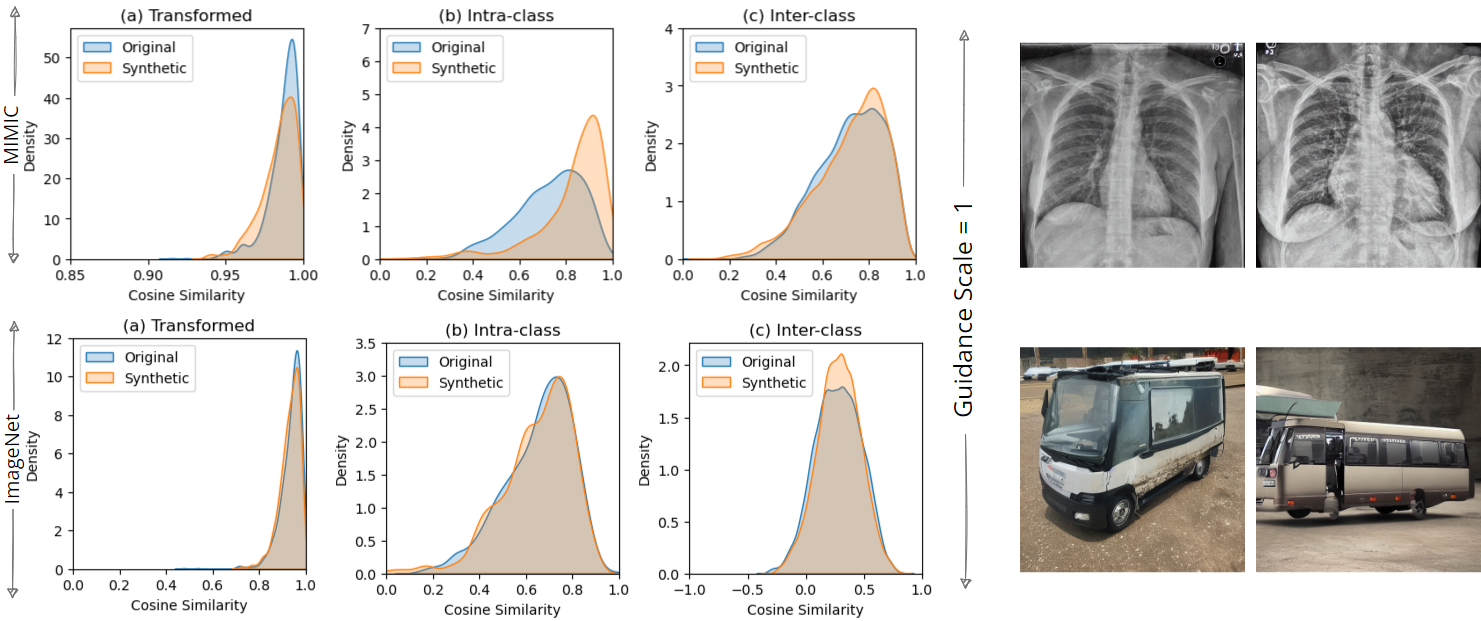}
    \includegraphics[width=0.9\textwidth]{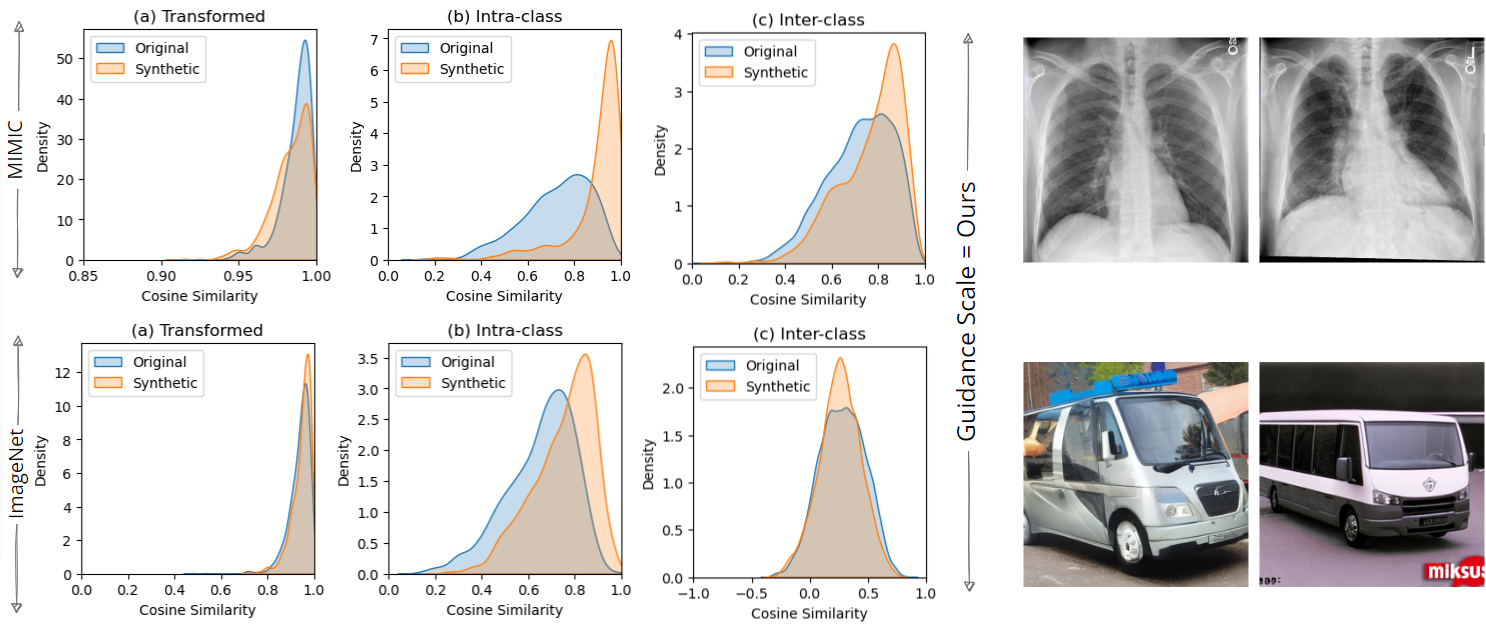}
    \includegraphics[width=0.9\textwidth]{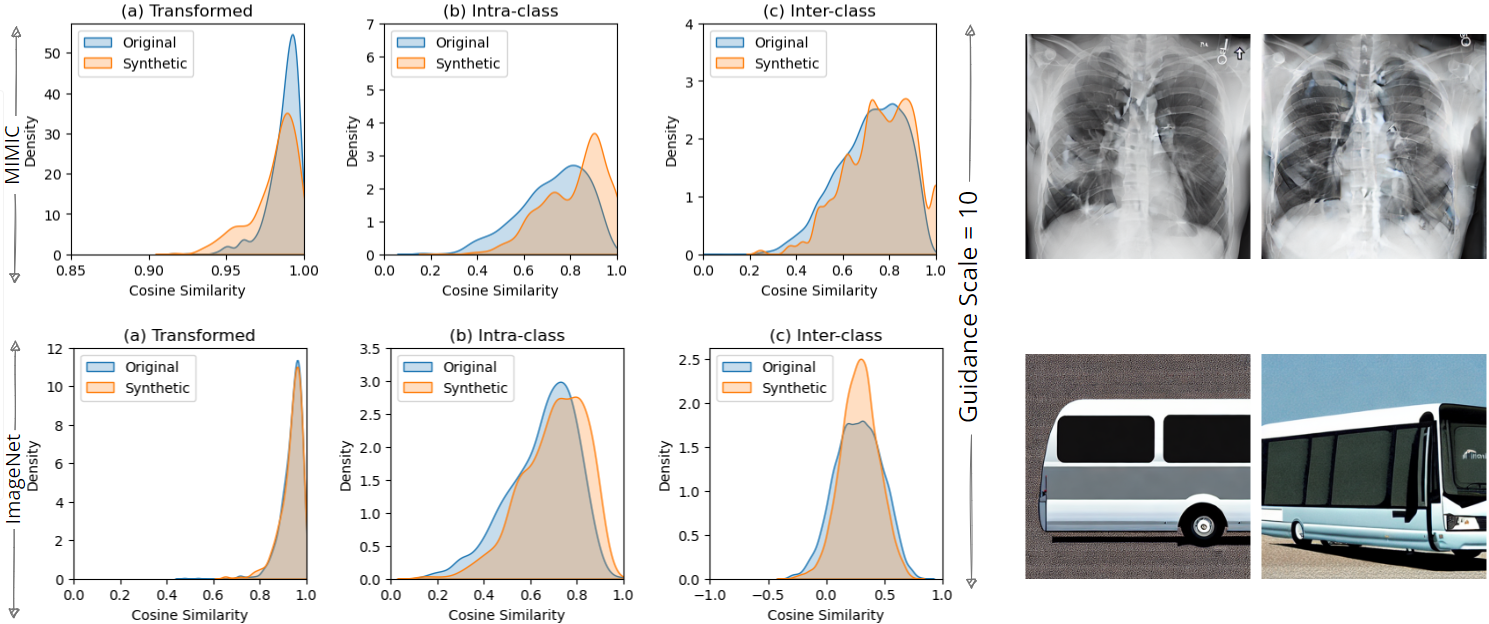}
    \vspace{10.5pt}
    \caption{Impact of Guidance Scale on Image Quality: Density plots comparing the distribution of real and synthetic images for transformed, intra-class, and inter-class scenarios. Three guidance scale values 1, 4 (ours), and 10 are explored for both Stable Diffusion and Roentgen models, revealing the influence of the guidance scale on the visual fidelity of synthetic images}
    \label{fig:kde_cfg_scales}
\end{figure*}

We explore the significance of guidance scale and its influence on the quality of synthetic images generated by Stable Diffusion and Roentgen models for both MIMIC-CXR and ImageNet datasets.  The guidance scale acts as a control parameter, influencing the fidelity of synthetic images which holds significance in the context of image generation models, serving as a guiding force in the synthesis process. In a technical sense, the guidance scale regulates the contribution of the guidance signal during the optimization process of the generative model. Its significance lies in guiding the generation process to strike a balance between realism and diversity. A carefully chosen guidance scale can enhance the quality and relevance of generated images to the target dataset. Density plots presented in Figure \ref{fig:kde_cfg_scales} visually assess the distribution of real and synthetic images across three scenarios: transformed, intra-class, and inter-class. They are accompanied by sets of synthetic samples corresponding to different values of the guidance scale. \\

We explore three scenarios: when the guidance scale is set to 1, resulting in low-quality images which deviate significantly from the realism of actual datasets for both Stable Diffusion and Roentgen models; the second scenario employs a guidance scale of 4, which we found to yield optimal results in our study; and the third scenario involves an increased guidance scale of 10. Upon visual inspection, we observe that a guidance scale of 4 strikes a balance, producing synthetic images that closely resemble real images. Deviating from this optimal value, such as increasing the scale to 10, leads to a degradation in image quality. The images generated under these conditions tend to exhibit exaggerated and unrealistic features, compromising the fidelity and utility of the synthetic data. These findings highlight the delicate balance required in selecting the guidance scale to achieve optimal results in synthetic image generation.
\label{sec:rationale}
\label{appA}

\subsection{Qualitative Diversity Assessment}
We conduct a visual assessment of images from the MIMIC-CXR and ImageNet datasets as well as their corresponding synthetic counterparts generated using the Roentgen and Stable Diffusion models, respectively. Specifically, we analyze the synthetic images corresponding to the most and least diverse classes identified using the proposed \texttt{SDICE} index. From Figure \ref{fig:classwise_box}, it can be observed that the `Atelectasis' is identified as the least diverse class, while `Fracture' is the most diverse class in CXR generation. Similarly, for natural image generation, `minibus' is the class with the least diversity and `fireboat' is the most diverse class identified using the \texttt{SDICE} index.\\

\begin{figure*}[h]
    \centering
    \includegraphics[width=\textwidth]{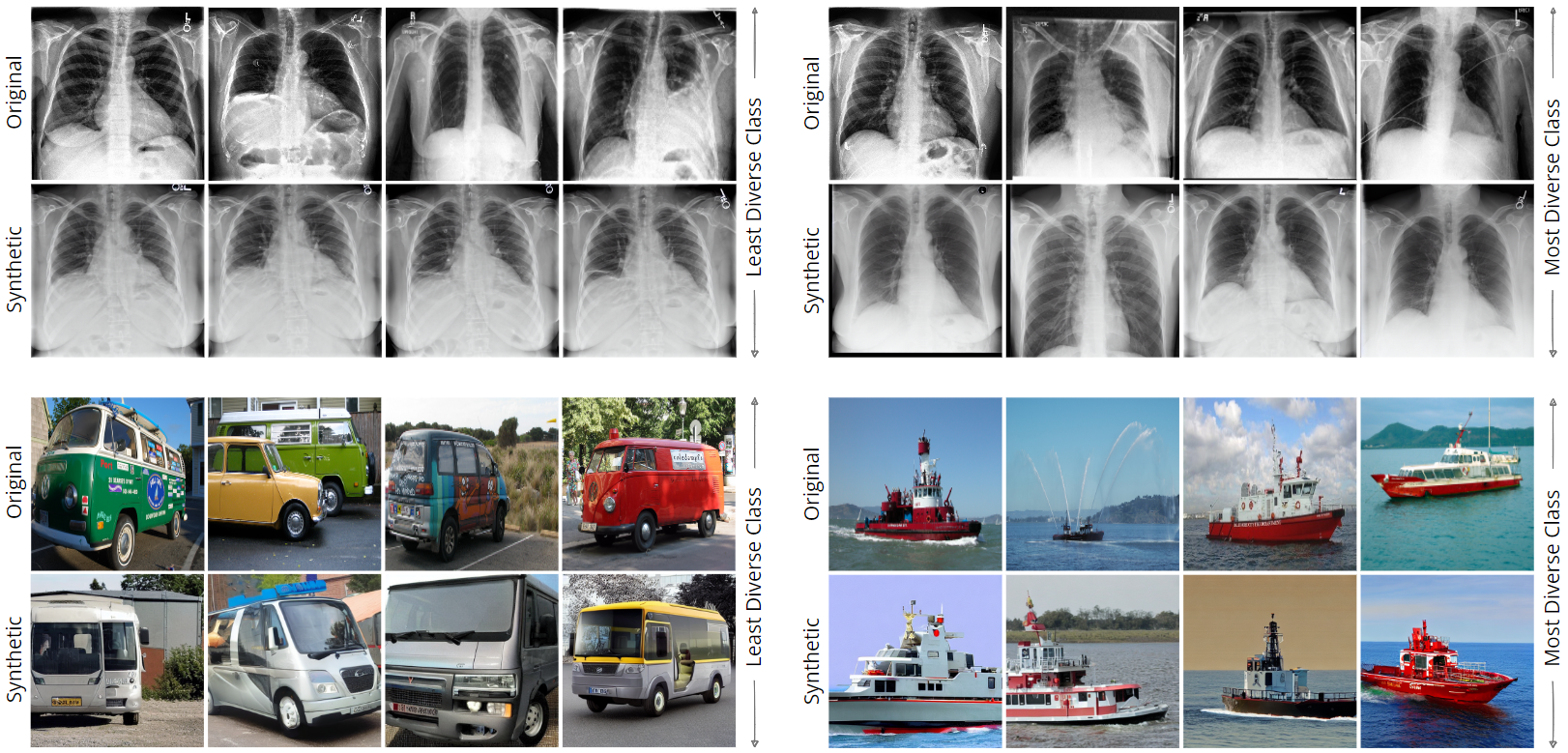}
    \vspace{4.0pt}
    \caption{Visual Comparison of Original and Synthetic Images: Examining diversity of CXR and natural synthetic images generated by the Roentgen and stable diffusion models, respectively. Only the images corresponding to the least and most diverse classes identified by the proposed \texttt{SDICE} index are shown here. Upper row illustrates samples from MIMIC-CXR, while lower row illustrates ImageNet samples.  `Atelectasis' in MIMIC-CXR exhibits the least diversity ($\gamma$=1.56e-08), while `Fracture' demonstrates the highest diversity ($\gamma$=7.93e-01). In ImageNet, `minibus' class has the least diversity ($\gamma$=5.52e-05), and `fireboat' stands out as the most diverse class ($\gamma$=9.97e-01).}
    \label{fig:supp_1}
\end{figure*}

From Figure \ref{fig:supp_1} it is obvious that all the synthetic images generated for the `Atelectasis' class look very similar, confirming their lower diversity. On the other hand, the synthetic images of the `Fracture' class exhibit reasonable diversity, which is correctly captured by the proposed \texttt{SDICE} index. Similarly, all the synthetic `minibus' images exhibit close similarity in color and shape unlike the corresponding real images, which justifies why they have a low \texttt{SDICE} index. 

\subsection{Catastrophic Forgetting Visualization}
\begin{figure}[htbp]
\centering
\begin{minipage}[t]{0.40\textwidth}
\centering
\includegraphics[width=\linewidth]{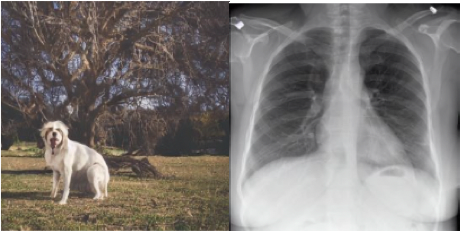}
% \caption{Prompt: "A Dog under a Tree"}
Prompt: "A Dog under a Tree"
\end{minipage}
\begin{minipage}[t]{0.40\textwidth}
\centering
\includegraphics[width=\linewidth]{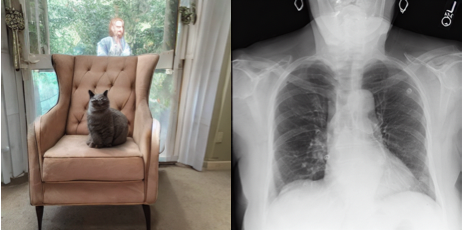}
% \caption{Prompt: "Cat on a Chair"}
Prompt: "Cat on a Chair"
\end{minipage}
\centering
\begin{minipage}[t]{0.40\textwidth}
\centering
\includegraphics[width=\linewidth]{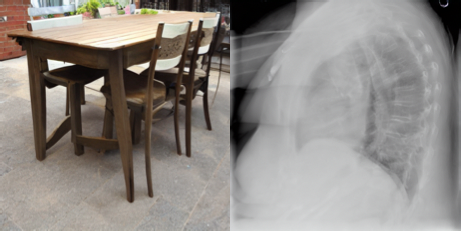}
% \caption{Prompt: "Table Chair"}
Prompt: "Table Chair"
\end{minipage}
\begin{minipage}[t]{0.40\textwidth}
\centering
\includegraphics[width=\linewidth]{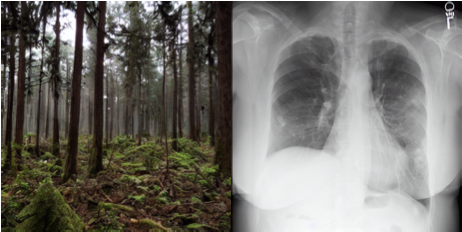}
% \caption{Prompt: "Dense Forest"}
Prompt: "Dense Forest"
\end{minipage}
\vspace{8.9pt}
\caption{Illustration of Catastrophic Forgetting observed in the RoentGen fine-tuned model. Image Generated when passed through a Diffusion Model (Left) and RoentGen Model (Right)}
\label{Figure 2}
\end{figure}

Catastrophic forgetting is a common problem in machine learning that occurs when a model forgets what it has previously learned when it is trained on new data. This is often the case when a model is fine-tuned on a specific domain, such as chest x-rays, and then trained on a different domain. In the case of the RoentGen model, the model was fine-tuned on chest x-rays to improve the overall performance. However, this fine-tuning process caused the model to forget what it had previously learned, and its performance suffered when it was used on new, unseen data outside of the chest x-ray domain. The authors noted that as the model was introduced to new images, its weights rapidly changed, leading to knowledge collapse. We experimented with this by passing different prompts such as "A Dog under a Tree", "Cat on a Chair", "Table Chair" and "Dense Forest" through the model, We noticed that the model still generated different chest x-rays. Figure \ref{Figure 2} shows the result on different prompts. We run this as a replication experiment and suspect that the RoentGen Model is susceptible to overfitting and may lack diversity or even memorize the training data.
% \vspace{-15.0}

\end{document}